\pdfoutput=1

\documentclass[11pt]{article}

\usepackage[]{acl}

\usepackage{tcolorbox}
\usepackage{times}
\usepackage{latexsym}
\usepackage{float}
\usepackage[T1]{fontenc}

\usepackage[utf8]{inputenc}

\usepackage{microtype}

\usepackage[utf8]{inputenc} 
\usepackage[T1]{fontenc}    
\usepackage{hyperref}       
\usepackage{url}            
\usepackage{booktabs}       
\usepackage{amsfonts}       
\usepackage{nicefrac}       
\usepackage{microtype}      
\usepackage{xcolor}         
\usepackage{subcaption}
\usepackage{caption}
\usepackage{wrapfig}
\usepackage{listings}
\usepackage{algorithm}
\usepackage{algpseudocode}
\usepackage{bbm}
\usepackage{soul}
\usepackage{graphicx}

\newcommand{\hlc}[2][yellow]{{%
    \colorlet{foo}{#1}%
    \sethlcolor{foo}\hl{#2}}%
}
\usepackage{multirow}
\usepackage{multicol}
\usepackage{makecell}
\usepackage{enumitem}

\usepackage[scaled=0.85]{beramono}            
\definecolor{color-blind-cyan}{HTML}{56B4E9}
\definecolor{color-blind-orange}{HTML}{E69F00}
\definecolor{color-blind-green}{HTML}{029E73}

\usepackage{amsmath,amsfonts,bm}
\usepackage{mathabx}

\def\eqref#1{equation~\ref{#1}}

\def\1{\bm{1}}

\DeclareMathAlphabet{\mathsfit}{\encodingdefault}{\sfdefault}{m}{sl}
\SetMathAlphabet{\mathsfit}{bold}{\encodingdefault}{\sfdefault}{bx}{n}

\newcommand{\orange}[1]{\textbf{\textcolor{color-blind-orange}{#1}}}
\newcommand{\green}[1]{\textbf{\textcolor{color-blind-green}{#1}}}

\newcommand{\diffdel}[1]{\colorlet{foo}{red!50}\sethlcolor{foo}\hl{#1}}
\newcommand{\diffadd}[1]{\colorlet{foo}{green!30}\sethlcolor{foo}\hl{#1}}

\newcommand{\cznote}[1]{\textcolor{blue}{}}

\newcommand{\memfree}[0]{\textsc{MemFree}}
\newcommand{\baseline}[0]{\textsc{Baseline}}

\usepackage{hyperref}
\hypersetup{
     colorlinks=true,
     urlcolor=blue,
     citecolor=black
}
\usepackage{url}
\usepackage{graphicx}

\makeatletter
\newcommand\footnoteref[1]{\protected@xdef\@thefnmark{\ref{#1}}\@footnotemark}
\makeatother

\title{Preventing Generation of Verbatim Memorization in Language\\ Models Gives a False Sense of Privacy}

\author{
Daphne Ippolito$^1$ \And Florian Tramèr\thanks{\,\,\,Remaining authors ordered by Algorithm \ref{alg:authorder} in Appendix \ref{authorder}; briefly, we require Daphne be listed first, and Nicholas listed last, and we search for the first permutation of authors' first names which satisfies these constraints, where permutations order names by their salted MD5 hash.}\,\,\,$^2$ \And Milad Nasr$^{*1}$ \AND Chiyuan Zhang$^{*1}$ \And Matthew Jagielski$^{*1}$ \And
Katherine Lee$^{*1,3}$ \AND Christopher A. Choquette-Choo$^{*1}$ \And Nicholas Carlini$^1$ \AND
{\normalfont \emph{$^1$ Google Research\qquad $^2$ ETH Zurich\qquad $^3$ Cornell University}}
}

\begin{document}

\maketitle

\begin{abstract}
Studying data memorization in neural language models helps us understand the risks (e.g., to privacy or copyright) associated with models regurgitating training data and aids in the development of countermeasures.
Many prior works---and some recently deployed defenses---focus on ``verbatim memorization'', defined as a model generation that exactly matches a substring from the training set.
We argue that verbatim memorization definitions are too restrictive and fail to capture more subtle forms of memorization.
Specifically, we design and implement an efficient defense that \emph{perfectly} prevents all verbatim memorization. 
And yet, we demonstrate that this ``perfect'' filter does not prevent the leakage of training data.
Indeed, it is easily circumvented by plausible and minimally modified ``style-transfer'' prompts---and in some cases even the non-modified original prompts---to extract memorized information.
We conclude by discussing potential alternative definitions and why defining memorization is a difficult yet crucial open question for neural language models.

\end{abstract}

\section{Introduction}

The ability of neural language models to memorize their training data has been studied extensively \citep{kandpal2022deduplicating,lee2021deduplicating,carlini2022quantifying,zhang2021counterfactual,thakkar2021understanding,ramaswamy2020training}.
When language models, especially ones used in production systems, are susceptible to \emph{data extraction} attacks, it can lead to practical problems ranging from privacy risks to copyright concerns.
For example, \citet{carlini2021extracting} showed that the GPT-2 language model could output personally identifying information of individuals contained in the training dataset.

One natural way to avoid this risk is to filter out any generations which copy long strings verbatim from the training set.
GitHub's Copilot, a language-model-based code assistant, deploys this defense by giving users the option to ``block suggestions matching public code''~\citep{githubcopilot}.

In this work, we ask the question:  ``\emph{Do language models emit paraphrased memorized content?}''
This scenario can happen  maliciously (e.g., adversaries trying to extract private user data) or through honest interactions (e.g., users prompting in real-world scenarios).
Indeed, we find that Copilot's filtering system is easy to circumvent by applying plausible ``style transfers'' to the prompt. For example, by translating variable names from English to French the model outputs completely memorized examples, but post-processed with the en-fr style transfer. 
We further show that GPT-3~\citep{brown2020language}, a model trained on natural language, is also vulnerable to extraction attacks.


\begin{figure}[t]
    \centering
    \includegraphics[width=\linewidth]{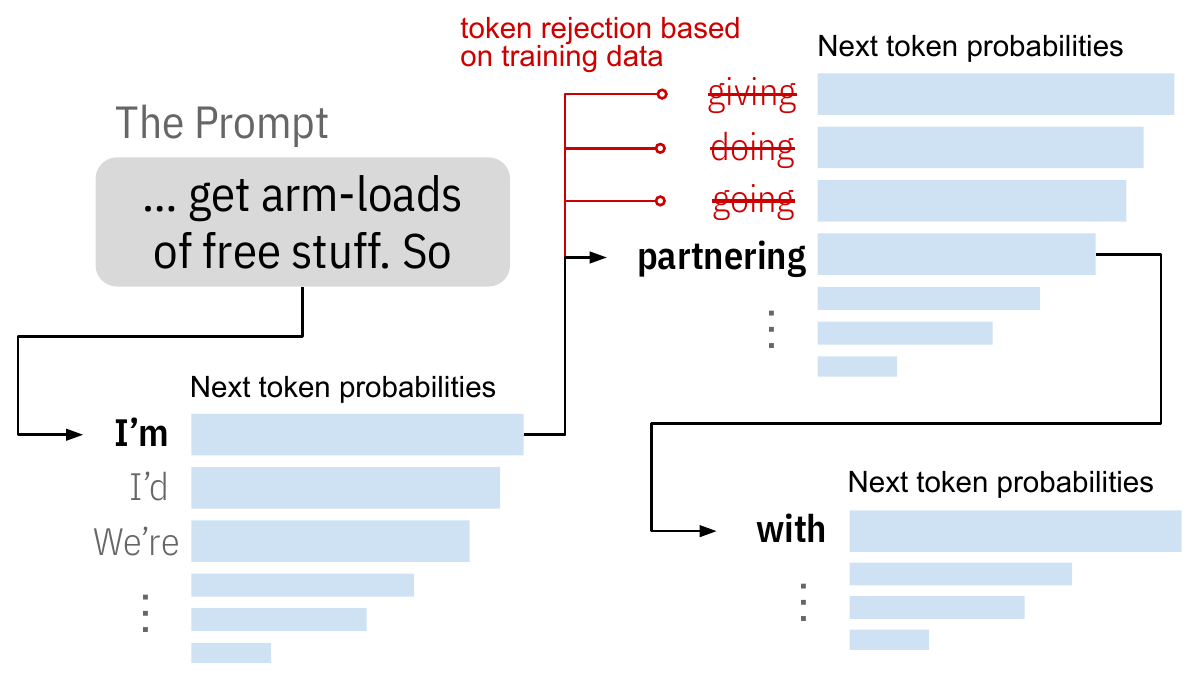}
    \caption{Illustration of Memorization-free Decoding, a defense which can eliminate verbatim memorization in the generations from a large neural language model, but does not prevent approximate memorization.}
    \vspace{-0.75em}
    \label{fig:bloom-illust}
\end{figure}

Unfortunately, Copilot's training set and precise algorithm for their defense are non-public.
Therefore, to investigate this phenomenon systematically, we develop \memfree{} decoding (Figure~\ref{fig:bloom-illust}), an efficient defense that is guaranteed to prevent all verbatim memorization, and which scales to training sets consisting of hundreds of gigabytes of text.
In \memfree{} decoding, at each step of generation we check whether the model's chosen next token would create an $n$-gram found in the training set.
If it does, an alternative next token is selected (without a computationally expensive regeneration) by sampling from the model's token posterior.
The check for membership in the training set is performed efficiently using a Bloom filter containing all common $n$-grams from the training set. 

We use \memfree{} to study Copilot's
verbatim-filtering defense on other state-of-the-art large language models such as GPT-Neo \citep{gao2020pile}. 
We first confirm that even honestly designed prompts often bypass verbatim memorization checks. Then, we observe another interesting phenomenon:
language models succeed at emitting \emph{approximate memorization} that bypass our filter all by themselves.
Indeed, when prevented from generating exact $n$-grams from the training set, models are capable of ``cheating'' the filter by producing close paraphrases--for example, inserting spelling errors, adjusting punctuation or whitespace, or using synonyms (e.g., swapping `and' with `\&').
These changes lead to generated text a human would perceive as 
nearly
identical, even if it is not verbatim memorization.


Clearly, defenses which prevent verbatim copying are \emph{necessary but not sufficient} to protect against training data leakage.
As a result of these failure modes, we argue that a broader definition of memorization is necessary when reasoning about training set memorization in language models.
Such a definition should not only capture verbatim notions of memorization, but also notions based on high ``semantic similarity'' between model outputs and training data. We conclude our work by comparing approximate and verbatim memorization, discussing their relation to other domains of literature, and the challenges surrounding the ambiguity of approximate memorizations.
%
%
Future work that aims to faithfully measure or prevent memorization in language models will need to take this ambiguity into account---for example, our analysis suggests that the fraction of datasets that large language models is likely far larger than the fraction as reported in prior work \cite{carlini2022quantifying}.

\section{Background}
\paragraph{Language Models.}
We consider auto-regressive language models that operate over sequences of text and, given a prefix $p$, output a probability distribution for the next token in the sequence.
To generate 
text for a prompt $p$,
the language model starts with an empty suffix $s$, and repeatedly samples the next token from its prediction on $p + s$, and then appends this token to $s$.
The success of neural language models has, in large part, been driven by the transformer architecture introduced of~\citet{vaswani2017attention},
which allowed models to scale from millions to hundreds of billions of parameters over the past half-decade~\citep{brown2020language,chowdhery2022palm,zhang2022opt}.
This increase in model sizes has likewise driven increases in dataset sizes, with most of this data coming from internet crawls~\citep{lee2021deduplicating,raffel2020exploring, gao2020pile}.\footnote{A common source for datasets is the Common Crawl dataset found at: \url{https://commoncrawl.org/}}


Prior work has shown that large language models can memorize and regurgitate potentially private information, like phone numbers and addresses, as well as memorize long sequences from their training sets~\citep{carlini2019secret,carlini2021extracting,lee2021deduplicating,carlini2022quantifying,zhang2021counterfactual,thakkar2021understanding,ramaswamy2020training,kandpal2022deduplicating}. Our work focuses on large language models trained to generate English text or code, and our work does not distinguish between problematic memorization (e.g. exposure of private information) and non-problematic memorization (e.g. quoting perfectly from a presidential speech).
%

\paragraph{Measuring Memorization.}
Many studies of memorization stem from a concern of privacy leakage:
if a model memorizes sensitive training data and can generate it, then interactions with a model can lead to the leakage of that sensitive data. Nearly all of this literature is focused on measuring \emph{verbatim cases of memorization}.

\emph{Eidetic memorization}~\citep{carlini2021extracting} defines a string $s$ as memorized if there exists a prompt $p$ so that
$\text{LM}(p) = s$ and $s$ is contained in the training dataset. This definition and variations of it have been used widely in the literature~\citep{kandpal2022deduplicating,lee2021deduplicating,carlini2022quantifying}. For example,~\citet{tirumala2022memorization} study a similar per-token definition called \emph{exact memorization} and~\citet{kandpal2022deduplicating} a document-level definition called \emph{perfect memorization}.





There is also a newly emerging line of works exploring \emph{differential-privacy (DP)-based definitions}~\citep{zhao2022provably,stock2022defending}, which relate to document-level DP guarantees in language modelling~\citep{yu2021differentially}. These works differ from the above in that they define a probabilistic leakage measure. However, this is based on the probability of generating---verbatim---a canary sentence $s$, depending on whether $s$ was contained in the training set or not. There are different probabilistic definitions, also based on verbatim sequences, such as the \emph{counterfactual memorization} proposed by~\citet{zhang2021counterfactual}. 

In the domain of language model memorization, the most similar work to ours is \citet{lee2021deduplicating} who also argue for a more relaxed definition of memorization. Lee \emph{et al.} say any model output for a prompt $p$ is memorized if it is within some chosen edit distance of the prompt's true continuation in the training set. As we will discuss, a small edit distance may not capture all forms of approximate memorization either---such as our examples of ``style-transfer'' applied to memorized content. 

\paragraph{Preventing Memorization.}
Differentially private training, e.g., using DP stochastic gradient descent~\citep{abadi2016deep}, is the gold standard for training models which provably do not memorize individual training examples.
However, in practice, these techniques result in worse generative models~\citep{anil2021large}---thus, no state-of-the-art, large, language models are trained with DP. Instead, data deduplication has arisen as a pragmatic countermeasure against data memorization~\citep{lee2021deduplicating,kandpal2022deduplicating, carlini2022quantifying}. The core idea is to remove any duplicated content---e.g., repeated documents---because duplicated content is much more likely to be memorized. However, deduplication does not \emph{guarantee} that a model will not still memorize individual (deduplicated) examples, necessitating defenses that operate at inference-time.

\section{Preventing Models from Emitting Verbatim Training Data}

In this paper, we consider inference-time defenses that eliminate the generation of memorized content from the training set.
The most immediate way to do this is 
simply to filter all model outputs using some fixed definition of memorization.
For example, in~\citet{carlini2022quantifying}, 
a continuation $s=\text{LM}(p)$ of a $k$-length prompt $p$ is said to be memorized if the string $s$ exists verbatim in the training dataset.
A straightforward implementation checks 
each generation $s$ against the training set and rejects any matches.
We call the approach of re-running a language model, possibly many times with different seeds, until a qualifying generation is produced, \textbf{retroactive censoring}.

The problem with retroactive censoring is that it effectively prevents the model from emitting any output when the model's confidence in a memorized string is too high. To encourage a model to generate novel outputs, we could also adopt a more granular filtering approach: rather than censoring memorized content solely at the level of an entire sequence $s$, we could instead check and mark each $n$-gram within $s$ individually.
Filtering for memorization at the $n$-gram-level rather than at the sequence level allows substrings of a generation which may be novel to be kept, and only the pieces that are verbatim memorized to be modified.
%
We call this approach \textbf{\memfree{} decoding}, as the defense is applied at decoding time.

Both retroactive censoring and \memfree{} decoding
explicitly prohibit the model from emitting a sequence if it is contained (entirely or partially) in the training dataset.
However, in retroactive censoring, if a generation starts off with memorized text, but then veers off track from the true continuation (a common occurrence), this would not be marked as memorization, even though a portion of the output sequence is clearly memorized.
The \memfree{} decoding approach performs a more fine-grained and aggressive check by filtering out all memorized subsequences of a given length.
In this work we use the \memfree{} decoding approach to show that even when a model is restricted from emitting any output with snippets of verbatim memorization, the model can still leak training data.



\subsection{\memfree{} Decoding Details}
In order to implement \memfree{} decoding, we alter the model's generation in an online manner by restricting the production of tokens which would result in an $n$-gram memorization. Let $p$ be the current working prefix and $t$ be the next proposed token when running the model forward.

Our algorithm first checks if 
any $n$-gram in the concatenated sequence $p || t$ is contained in the training dataset $D$. If it is, we suppress this generated token and re-sample from the model. 
To avoid potentially expensive resamplings, we equivalently express this as altering the model's output probability distribution by removing the probability mass from token $t$. In this way, we guarantee that prior to sampling the probability of outputting a memorization will be $0$.
Appendix \ref{section:algorithm} gives a formal procedure for this method. 

%
%
%
%
Altering the token posterior allows any sampling strategy to be used on top of memorization-free decoding.
For example, if one uses top-$k$ sampling, tokens that result in memorization are disqualified before the probability distribution is truncated to the $k$ next most likely tokens.
This procedure is guaranteed to generate non-memorized text.

\subsection{Querying the Training Set Efficiently}
\label{ssec:memorization-check-efficient}
Our \memfree{} defense has assumed that it is easy to perform the query $s \in D$ to test if any given
string is contained in the training dataset. 
Because the defense works at inference-time, it is necessary that this query is computationally efficient to maintain utility of the language model.
Given that training sets may contain terabytes of data \citep{brown2020language}, it is infeasible to maintain an entire copy of the training dataset in an efficiently accessible storage.
Thus, we explore three optimizations to speed up the process of memorization checking.

\textbf{First}, as a direct result of our $n$-gram memorization definition, we can equivalently check only the $n$-gram ending in the current predicted token $t$; we can thus avoid many $n$-gram queries for each token. Further, and in addition to preventing subsequence memorization, this allows us to avoid queries into a large set of all prefixes and continuations.

\textbf{Second}, we only check against sequences that have a reasonable probability of being memorized by the model.
In theory, this could be easily determined by running each 
$n$-gram $s \in \mathcal{D}$ through the model and then filtering out all sequences with high loss (thus unlikely to be memorized). However, 
this is a computationally
expensive procedure as it requires re-processing every substring of the training dataset.
Instead, a computationally- and storage-efficient procedure could be to only store $n$-grams which occur more than once in the training set---prior work has shown duplicate text is the most likely to be memorized~\citep{lee2021deduplicating,kandpal2022deduplicating}.

\textbf{Third}, by being willing to tolerate some false positives (labeling an $n$-gram as memorized when it is in fact not), we can take advantage of probabilistic data structures such as Bloom filters \citep{bloom1970space}, which admits no false negatives but trades off time and space with the false positive rate (which can be computed exactly).
Thus, by using a Bloom Filter, we guarantee that no memorized $n$-gram will ever be released (i.e., a false negative) but we may (rarely) prevent the emission of non-memorized content (i.e., a false positive).

Integrating a Bloom Filter into our defense is straightforward. Let $\mathcal{F}_{fp}(\mathcal{D}_n)$ represent the Bloom Filter of dataset $\mathcal{D}$, generated by adding each $n$-gram of the dataset $s \in \mathcal{D}_n$ to the Bloom filter, with false positive rate $fp$. Then, any memorization check $s \in \mathcal{D}_n$ in Algorithm~\ref{alg:memfree_decoding_algorithm} can be replaced with $s \in \mathcal{F}_{fp}(\mathcal{D}_n)$. The Bloom filter can be generated with a single pass over the model's training set, which could be performed in parallel with one epoch of model training. 

\paragraph{Additional Parameters.}
We must choose an appropriate false positive rate based on memory constraints and the chosen $n$-gram length. Choosing $n$ has two major impacts: on the population size (i.e., the number of unique $n$-grams) and thus the size of the filter, and on the effectiveness of memorization mitigation. 
If $n$ is set too low, then we will certainly prevent all memorized sequences but might also prevent too many common phrases.
But if we set $n$ too high, we might not prevent actually memorized sequences from being emitted by the model.
We discuss these tensions in Appendix~\ref{app:bloom}, along with two additional takeaways: (1) that \memfree{} \emph{does not impact downstream model performance} (which may result from false positives), and (2) that our chosen optimizations maintain a suitably low false negative rate (we observed a $3000$x improvement). These optimizations led to a filter of size $1.6$ gigabytes (or, $40.5$ gigabytes if all, even non-duplicated, $10$-grams were stored) when run over the 800GB Pile dataset.

\begin{figure}[t]
\footnotesize
\begin{subfigure}[t]{\linewidth}
{\small\textbf{Standard prompting with original prefix and format}}
\begin{lstlisting}[breaklines,frame=single,basicstyle=\tiny]
<@\textcolor{color-blind-cyan}{\texttt{float Q\_rsqrt( float number )}}@>
{
  long i;
  float x2, y;
  const float threehalfs = 1.5F;

  x2 = number * 0.5F;
  y  = number;
  i  = * ( long * ) &y;   
<@\textcolor{color-blind-orange}{Copilot no longer generates continuations}@>
\end{lstlisting}
\end{subfigure}
\begin{subfigure}[t]{\linewidth}
{\small\textbf{Prompt with Python-style comment}}
\begin{lstlisting}[breaklines,frame=single,basicstyle=\tiny,commentstyle=\ttfamily]
<@\textcolor{color-blind-cyan}{\texttt{\# float Q\_rsqrt( float number )}}@>
# {
#   long i;
#   float x2, y;
#   const float threehalfs = 1.5F;
#
#   x2 = number * 0.5F;
#   y  = number;
#   i  = * ( long * ) &y;        
#   i  = 0x5f3759df - ( i >> 1 );
#   y  = * ( float * ) &i;
#   y  = y * (threehalfs - (x2*y*y));
#
#   return y;
# }
\end{lstlisting}
\end{subfigure}
\begin{subfigure}[t]{\linewidth}
{\small\textbf{Prompt with French translation (alternate naming\\convention)}}
\begin{lstlisting}[breaklines,frame=single,basicstyle=\tiny]
<@\textcolor{color-blind-cyan}{\texttt{float Q\_sqrt( float nombre )}}@>
{
  long i;
  float x2, y;
  const float trois_moitie = 1.5F;

  x2 = nombre * 0.5F;
  y  = nombre;
  i  = * ( long * ) &y;
  i  = 0x5f3759df - ( i >> 1 ) 
  y  = * ( float * ) &i;
  y  = y * (trois_moitie - (x2*y*y)); 
  //y  = y * (trois_moitie - (x2*y*y));

  return nombre * y;
}
\end{lstlisting}
\end{subfigure}
    \caption{\textbf{Honest ``style-transfer'' prompts evade verbatim memorization filters.} Trivially modifying prompts causes GitHub's Copilot language model 
    to emit memorized, but not verbatim, content. 
    \textcolor{color-blind-cyan}{Prompts highlighted in blue.}  
    Model evaluated with the option ``block suggestions matching public code'' enabled. 
    For brevity, we removed comments from model outputs.}
    \label{fig:co-pilot-prompting}
\end{figure}

%

\subsection{Measuring Approximate Memorization}
\label{sec:def-approx-mem}
To show that defenses against verbatim memorization still allow approximate memorization, we need a definition for approximate memorization.
We consider two definitions.
First, drawing from standard NLP evaluation techniques, we measure the BLEU score between the generated and ground-truth continuations.
Second, we measure the length-normalized character-level Levenshtein similarity between the generated and ground-truth continuations.
Appendix \ref{app:eval_metrics_details} gives implementation details. 
In Section \ref{sec:experiments}, we investigates how these two similarity metrics decrease with \memfree{} decoding.

For situations requiring a binary label of whether approximate memorization has occurred, we use the following definition: a suffix $s$ for prefix $p$ is labeled as memorized if for generation $g=f(p)$,
$\text{BLEU}(g, s) > 0.75$.
This threshold was chosen by qualitatively inspecting examples.
Several example generations that are close to this threshold are shown in Table A\ref{tab:approx_mem_at_threshold}.

\begin{figure}[t]
    \centering
    \includegraphics[width=0.8\linewidth]{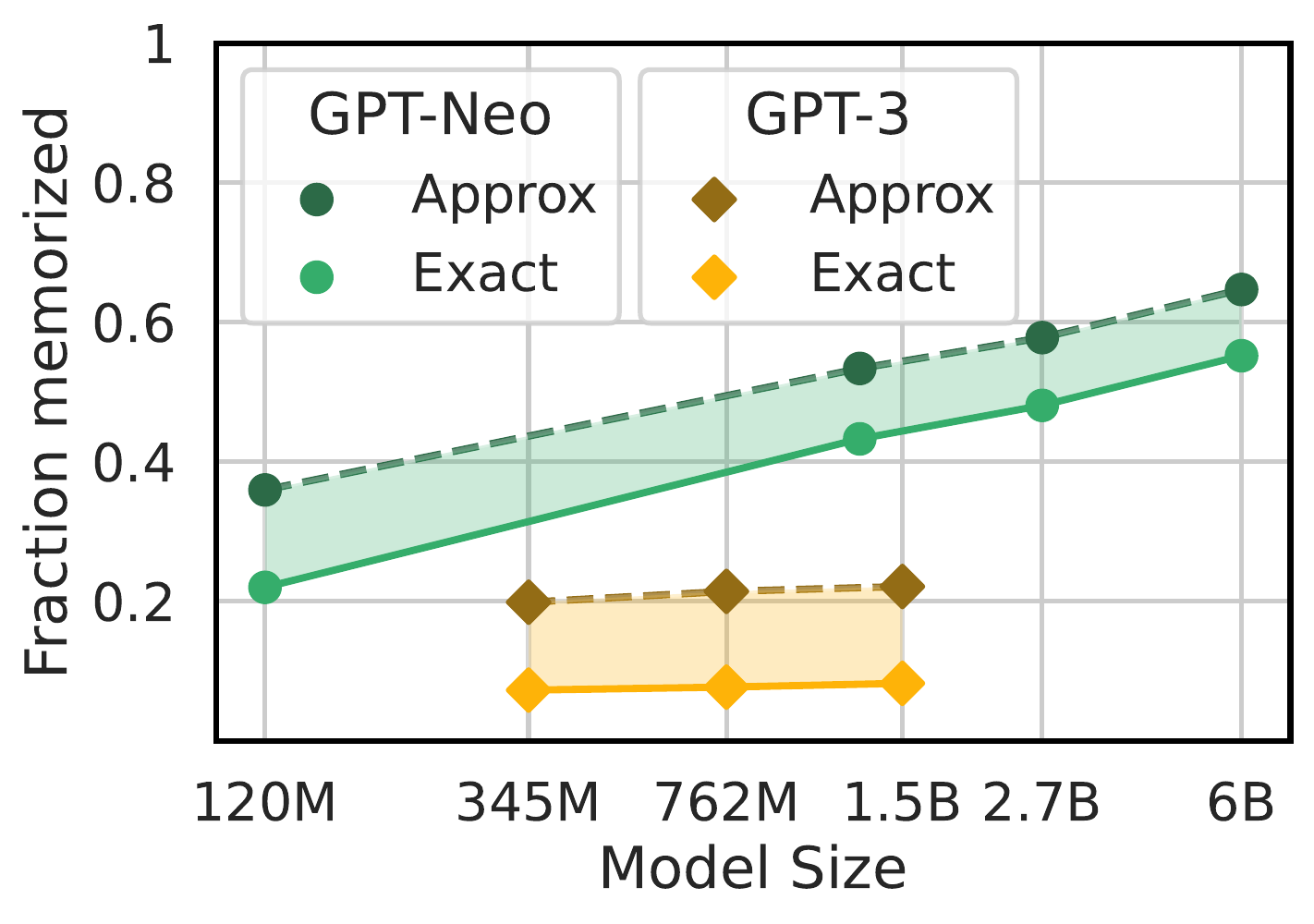}
    \caption{\textbf{Significantly more examples are approximately memorized (BLEU > 0.75) than are found to be exactly memorized by \citet{carlini2022quantifying}}. This is for undefended generation.}
    \label{fig:approx-scale}
 \end{figure}

When we repeat the prefix-extraction experiment from~\cite{carlini2022quantifying} to measure incidents of generations that could be considered memorized, but using this approximate definition instead of a verbatim one, we find that hat prior literature has significantly underestimated memorization leakage.
In Figure~\ref{fig:approx-scale}, the shaded region represents the fraction of memorized samples that would have bypassed a verbatim memorization filter: in the worst case, there is a factor-of-two increase. 

However, we caution that this definition of approximate memorization is inaccurate, potentially both over and under counting approximate memorization.  
While our choice of a $0.75$ BLEU score threshold shows a significant increase in approximate vs. verbatim memorization, it is not clear that all identified cases of memorization would be perceptually tagged as such by a human judge. 
This is one reason why simply switching to this definition for defenses may not be ideal---it could introduce significant false positives.

\section{Evading Verbatim Memorization Defenses}
\label{copilot}
In this section, we show how retroactive censoring of verbatim memorization can be evaded, even in settings where models are used honestly.
We first present a case study with Copilot, which has implemented retroactive censoring in production.
We then show how a large English language models like GPT-3 and PaLM are susceptible to the same vulnerability, should a defense similar to Copilot's be deployed.
In short, protecting against verbatim memorization can lead to a false sense of privacy.

\subsection{Evading Copilot's Memorization Filter}
Copilot is a code auto-complete service which is trained on GitHub code. Copilot is built using the Codex language model designed by OpenAI~\citep{chen2021evaluating}. To prevent generating memorized code, Copilot uses a filtering mechanism that blocks model outputs from being suggested if they overlap significantly (approximately 150 characters) with a training example. This is a practical example of a filter that aims at preventing perfect verbatim memorization, presumably by using a procedure similar to Algorithm~\ref{alg:memfree_decoding_algorithm} (the exact mechanism used by GitHub is not public).
However, we find that the filter fails to prevent the leakage of training data in many settings.

\paragraph{Style-transfer prompting.}
In Figure~\ref{fig:co-pilot-prompting}, we show that Copilot's filter can easily be bypassed by prompts that apply various forms of ``style-transfer'' to model outputs, thereby causing the model to produce memorized (but not verbatim) outputs. 

As a concrete example, we demonstrate how to extract the public code for Quake's ``Fast Inverse Square Root''.
If we naively prompt the model with the function definition \texttt{``float Q\_rsqrt ( float number )''},
Copilot correctly aborts generation of the full function (``standard prompting'').

However, we find that simple style-transfers applied to the prompt allow us to easily bypass Copilot's restrictions.
First, via prompting with ``Python-style comments'' we begin our prompt with Python's comment character ``\#''. Even though this is syntactically invalid C code, Copilot outputs the entire verbatim fast inverse square root algorithm, but commented out.
Second, in prompting with ``French translations'' we change the naming convention to French. 
As a result, the generations follow the new naming convention and are no longer flagged as a verbatim match. Other naming conventions, such as pre-pending ``$\_$'' to the variable or changing the language to Spanish, also work.


These strategies work because the Copilot model is sufficiently powerful: it can both follow the style-transfer prompt (by e.g., 
renaming variables) while simultaneously regurgitating memorized training data. We provide more examples in Appendix~\ref{app:examples-bypassing-copilot}.

\paragraph{Copilot evades its own filter.}
Not only do \emph{actively} style-transfered prompts evade the verbatim memorization filter, but even passively
prompting Copilot with highly duplicated text from the Pile dataset can too.
We find several examples where Copilot evades its own filter to output memorized text, some of which we show in Figure~\ref{fig:non_verbatim}. 
We see that Copilot evades the filter by (1) changing capitalization, (2) making small non-stylistic errors, and (3) changing whitespaces. The latter evasion (changing whitespaces) is surprising, as Copilot's documentation reports ignoring whitespace in its filtering mechanism (Appendix~\ref{app:co-pilot-filter}). However, we hypothesize that this can be explained by the model replacing tabs with space characters. We can verify this by adding tabs to the beginning of each line of the Q\_sqrt function, as an application of our style-transfer strategy.

\subsection{English Language Models} Following our analysis of Copilot, we ask whether this vulnerability is pervasive in other language models too.
We use API access to four large (English) language models---GPT-3 Davinci Original and V2 and PaLM 62B and 540B--to test whether they would be susceptible to style transfer of the prompt.
We assume that the training sets for these models are unknown and prompt with documents we believe are likely to have been memorized:
open-source licenses, famous speeches and monologues, novel openings, and song lyrics. 
For each document, we prompt the model with 100 words of either (1) the original document (``base''), (2) the document with all spaces doubled (``spaces''), (3) the document in all lowercase (``lower''), and (4) the document in all uppercase (``caps'').
We report approximate memorization results of this experiment in Figure~\ref{fig:famous-datasets}, with additional figures in Appendix~\ref{app:sec:large-english-model}.

\begin{figure}[t]
    \centering
    \includegraphics[width=\linewidth]{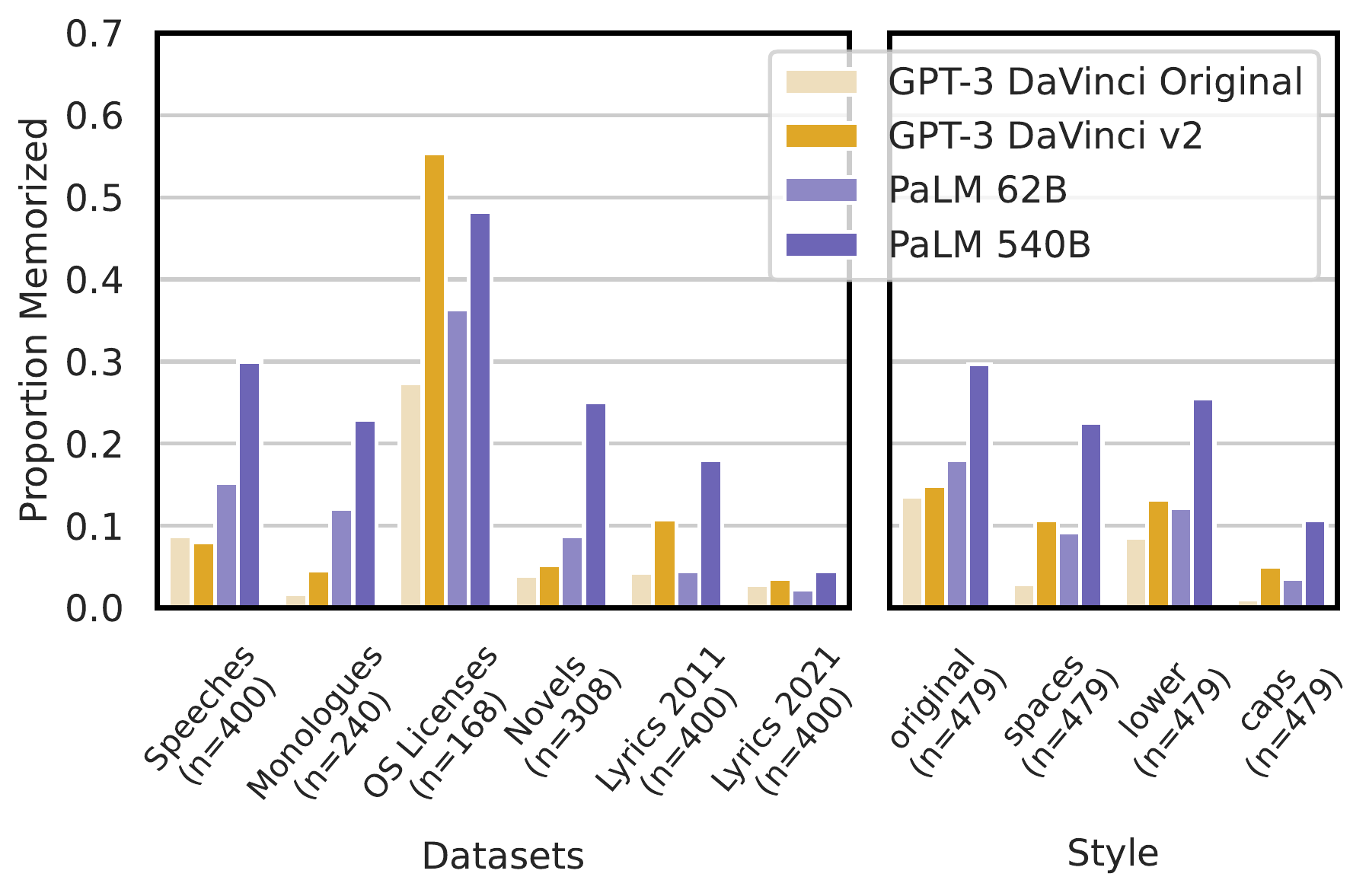}
    \caption{Fraction of prompts which discover approximate memorization, grouped by domain (left) and by style transfer applied (right).
    We tested two versions of GPT-3 DaVinci and two sizes of PaLM.
    Full plot in Appendix \ref{app:sec:large-english-model}.}
    \label{fig:famous-datasets}
 \end{figure}

We see that even when prompting with style-transfered prompts, GPT-3 and PaLM are still often able to generate memorized continuations.
Defenses for verbatim memorization are therefore incomplete.
Among the three techniques, uppercasing was the least likely to lead to memorized generations.
For the two PaLM models, the larger one is much more capable of memorization than the smaller one, which validates prior work \citep{carlini2022quantifying}.
The two version GPT-3 are purportedly the same size model but have quite different memorization tendencies.
For example, V2 is much more susceptible to the ``double spaces'' style transfer than the Original Davinci.
This emphasizes the importance of models' training set compositions and training methods on memorization tendencies.


\begin{figure}[t]
    \centering
    \begin{subfigure}[t]{\linewidth}
    {\small\textbf{1) Misspelling and changed capitalization}}
    \begin{lstlisting}[breaklines,frame=single,basicstyle=\tiny,breakindent=0pt,showstringspaces=false]
<@\textcolor{color-blind-cyan}{This program is free software}@>"; you can <@\green{redistribute}\orange{redistribut}@> it and/or * 
modify it under the terms of the GNU General Public License * <@\green{as}\orange{AS}@> published by the Free Software Foundation; either version 2 * of the License, <@\green{or}\orange{OR}@> (at your option) any later version"
    \end{lstlisting}
    \end{subfigure}
    \begin{subfigure}[t]{\linewidth}
    {\small\textbf{2) Small non-stylistic errors}}
    \begin{lstlisting}[breaklines,frame=single,basicstyle=\tiny,breakindent=0pt]
<@\textcolor{color-blind-cyan}{@aws-sdk/protocol-http"}@>;
<@\textcolor{color-blind-cyan}{import \{ Command as \$Command \} from "@aws-sdk/smithy-client";}@>
<@\textcolor{color-blind-cyan}{import \{FinalizeHandlerArguments,Handler,HandlerExecutionContext,}@>
<@\textcolor{color-blind-cyan}{MiddlewareStack,
   HttpHandlerOptions\} as \_\_HttpHandlerOptions,}@>MetadataBearer 
as __<@\green{MetadataBearer}\orange{MetadataBear}@>, 
    \end{lstlisting}
    \end{subfigure}
    \begin{subfigure}[t]{\linewidth}
    {\small\textbf{3) Changed whitespace}}
    \begin{lstlisting}[breaklines,frame=single,basicstyle=\tiny,breakindent=0pt]
<@\textcolor{color-blind-cyan}{[...] 
IPV6\_2292PKTINFO(2 ws)\textbackslash t\textbackslash t\textbackslash t\textbackslash t\textbackslash t = 0x2\textbackslash n}@><@\textcolor{color-blind-cyan}{IPV6\_2292PKTOPTIONS \textbackslash t\textbackslash t\textbackslash t\textbackslash t = 0x6\textbackslash n}@>
IPV6_2292RTHDR <@\green{[\textit{20 spaces}]}\orange{[\textit{9 spaces}]}@>= 0x5\n
    \end{lstlisting}
    \end{subfigure}
    \caption{\textbf{CoPilot can ``cheat'' and emit nearly verbatim memorized content.} Here, we show prompts from the training set, where the model makes slight errors causing the continuations to pass the filter. \textcolor{color-blind-cyan}{Prompts are in cyan}, followed by CoPilot's continuation where errors are highlighted as
    \orange{model's generation in orange} with the \green{correct characters in green}. 
    }
    \label{fig:non_verbatim}
\end{figure}

\section{\memfree{} Decoding Experiments}
\label{sec:experiments}
In this section, we 
study the effectiveness of our proposed \memfree{} decoding defense from Section~\ref{sec:def-approx-mem},
and the appropriateness of our proposed definition of approximate memorization.

\subsection{Experimental Design}
It is not possible to apply \memfree{} to the models from Section \ref{copilot} since their training sets are non-public.
Instead, we turn to the GPT-Neo languge model family~\citep{black2021gpt}.
These models are trained on the Pile, a publicly available 825GB dataset~\citep{gao2020pile}.
We build a Bloom filter over all $10$-grams occur 10 or more times.\footnote{Note that the choice of $n$=10 for the $n$-gram size is very conservative, and common phrases that happen to be composed of 10+ tokens will get filtered out by this check. We discuss why we chose these particular values in Appendix \ref{app:bloom}.}
In all experiments, we generate text using $\arg \max$ decoding as the sampling method.
We investigate four model sizes: 125M--6B parameters. 

We evaluate using substrings of the Pile released by \citet{carlini2022quantifying}.
The dataset includes 30k strings of length 150 tokens taken from the training set.
These are divided into 30 buckets of 1k strings, sampled such that the strings in bucket $i$ occur in the Pile between $2^{i/4}$ and $2^{(i+1)/4}$ times.
For each string, we use the first 50 tokens as a prompt $p$ and generate a 50-token long continuation.


\subsection{Reduction in Memorization} 
\begin{figure}[t]
     \begin{subfigure}[t]{0.23\textwidth}
          \centering
          
          \includegraphics[width=\linewidth]{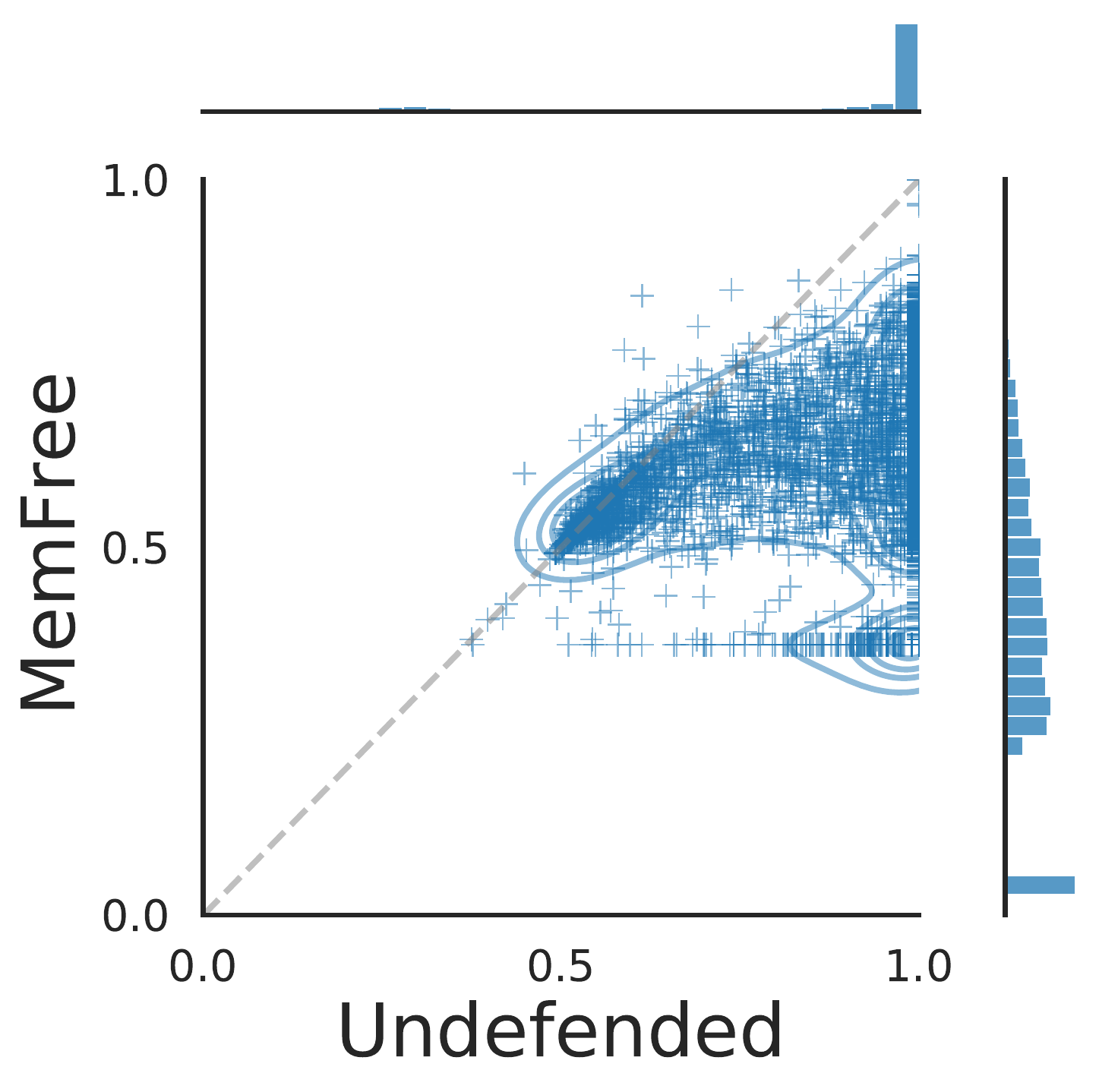}
          \caption{BLEU (word-level)}
          \label{fig:bleu_similarity}
      \end{subfigure}
      \hfill
      \begin{subfigure}[t]{0.23\textwidth}
          \centering
          \includegraphics[width=\linewidth]{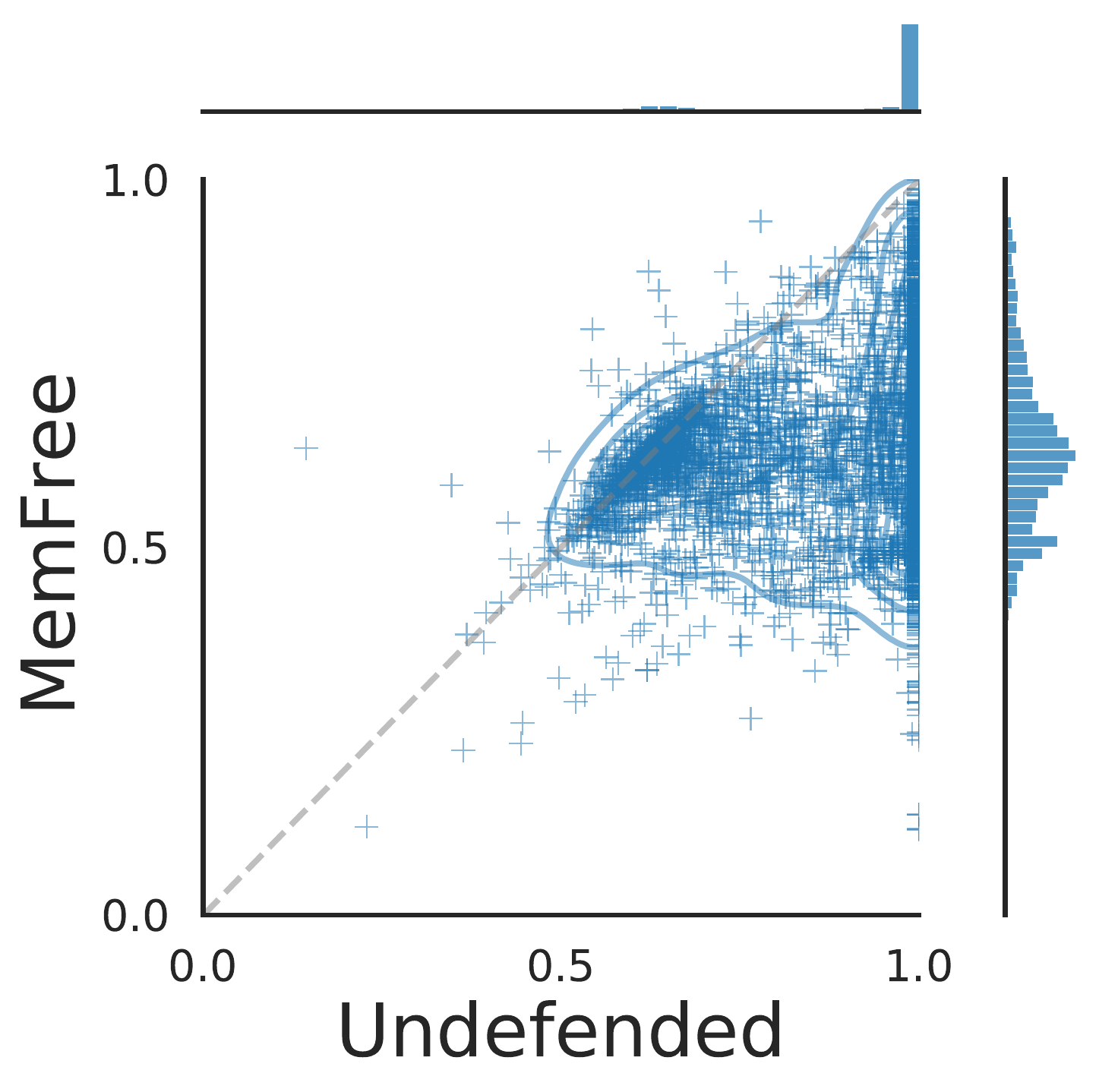}
          \caption{Edit similarity (char-level)}
          \label{fig:char_distance}
     \end{subfigure}
    \centering
    \caption{
    \textbf{\memfree{} reduces similarity when the continuation would have been highly similar to the ground-truth}, and has little impact otherwise.
     For 5,000 prompts, we plot the similarity of the groundtruth continuation with the generation from \memfree{} (y-axis) and with the undefended generation (x-axis).
     Generations on the diagonal were not memorized.}
     \label{fig:distance_scatter}

\end{figure}

\memfree{} significantly reduces the similarity of generations to the groundtruth, compared to performing undefended generation (Figure~\ref{fig:distance_scatter}). 
We also observe that when undefended generation already results in low similarity with the groundtruth, \memfree{} does not significantly alter the generations, as desired.

\begin{figure}[t]
    \centering
    \includegraphics[width=\linewidth]{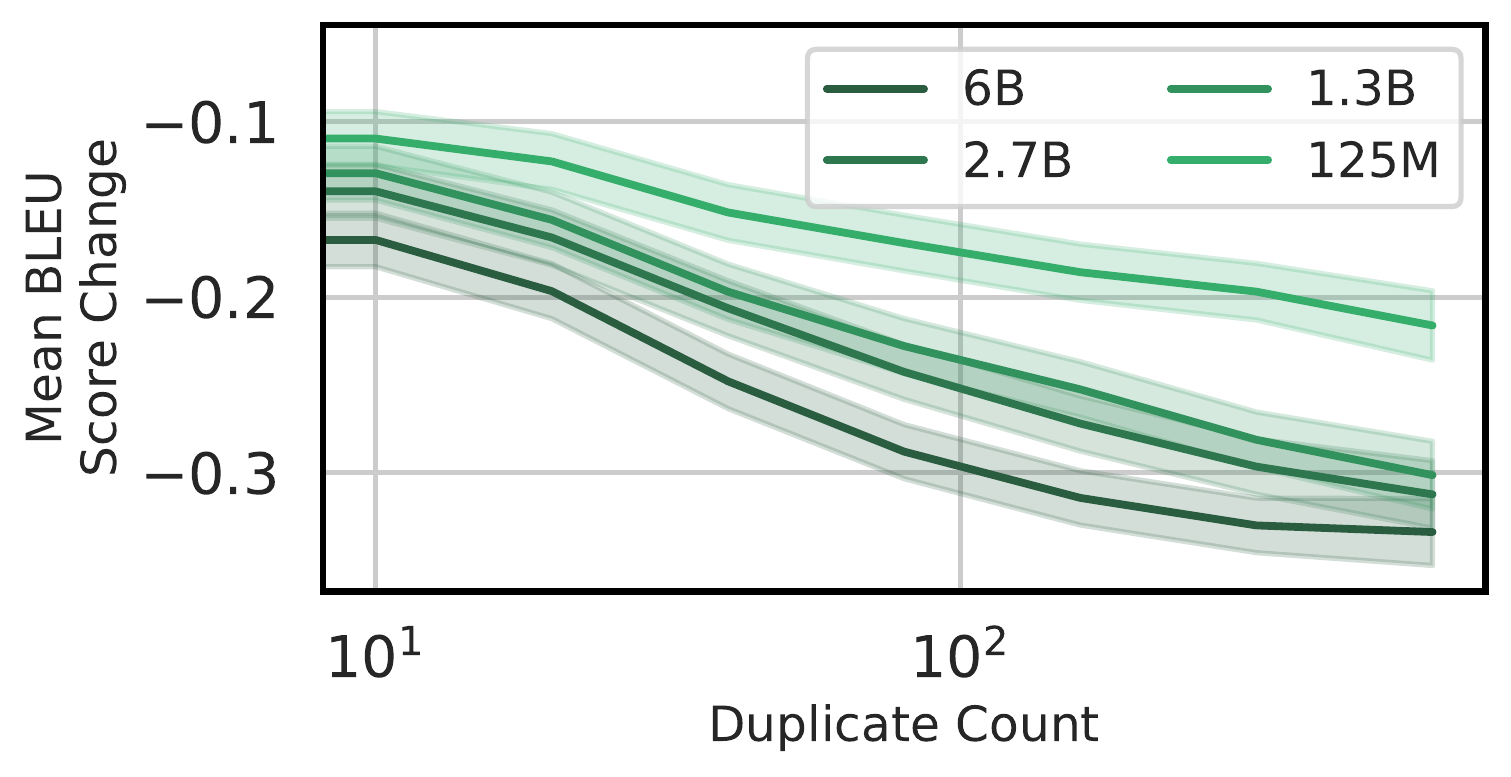}
    \caption{\textbf{\memfree{} decreases the BLEU score of generations more for highly duplicated examples}.}
    \label{fig:edit-dups}
 \end{figure}

\begin{figure}[t]
   \centering
          \includegraphics[width=\linewidth]{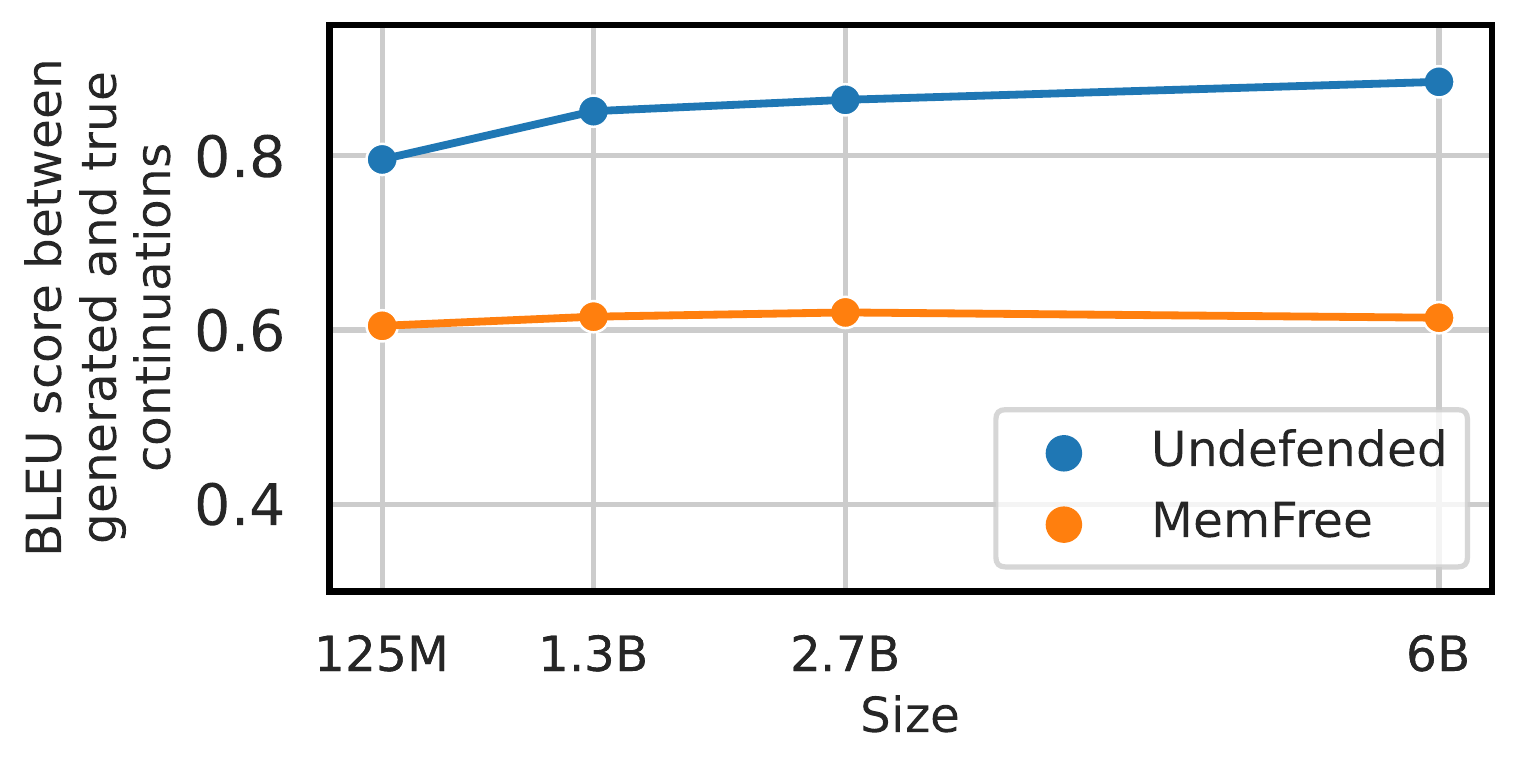}
        \caption{\textbf{\memfree{} remains effective at reducing similarity between the generated and groundtruth continuations even as models grow larger.},
        }
        \label{fig:modelsize}
\end{figure}

Previous work shows that increasing model size increases discoverable memorization ~\citep{carlini2022quantifying,kandpal2022deduplicating}.
We again find a clear trend that generations from larger models have, on average, a much higher similarity with the original continuation (Figure \ref{fig:modelsize}).
Despite this, \memfree{} remains effective at all model sizes (BLEU remains near-flat around $0.6$). 
Even when a sequence has many duplicates in the train set (a strong indicator of memorization), \memfree{} significantly decreases similarity with the groundtruth at all model sizes (Figure~\ref{fig:edit-dups}).


\subsection{Failures in Preventing Memorization}
A defense against memorization fails when it allows a sequence to be generated which a human would perceive as substantially copied from the true continuation---even if it is not verbatim memorized.
This failure case can be seen as the points where the \memfree{} generation is still a close match to the ground-truth continuation (Figure \ref{fig:distance_scatter}).
It occurs because the defense only adjusted a few tokens (e.g., $1$ after every sequence of $10$).
When looking at these examples, many, but not all, are lists of numbers.
Some examples are included in Table A\ref{tab:still_memorized}.
There is also a second failure-case: when a full ($50$ token) generation is made more similar with the ground-truth by \memfree{} (on $10$-grams) than without. This may happen depending on the model's token posterior's after removing all tokens that fail the \memfree{} check. Almost all of these cases had a trivial increase in similarity. However, $0.16\%$ of samples had a similarity increase above $0.1$.
We found qualitatively that many of these cases did have significant overlap with the true continuation.

\section{Discussion}

\paragraph{Defining memorization in language models.}
While verbatim definitions have helped discover significant memorization in large language models, 
they are insufficient to capture more subtle forms of memorization. 
Our work highlights two such situations: "style-transfer" prompting, where defenses for verbatim memorization can be actively subverted, and when models ``cheat'' by outputting similar, but not verbatim, continuations. As a result, our work suggests that memorization prevention must capture these types of paraphrased memorizations in addition to the previously considered verbatim definitions. 
However, exhaustively anticipating styles to incorporate into defenses is an innumerable problem that will become harder as models become more powerful. 

This emphasizes two major challenges in defining approximate memorization. 
First, since new approximate cases must be discoverable by the definition, this can result in a cat-and-mouse game.
Second, the definition of memorization
is domain-dependent.
For example, our paper focuses on language models trained to output English and code, which each have different standards for what it means to memorize.
Other languages will require different considerations when defining memorization.

There are a few areas of research which may help in improving memorization definitions. The field of \textbf{image generation memorization} is already comfortable with measuring \textit{fuzzy} (in our terms, \textit{approximate}) memorization, where generated items may be perceptually similar to training set examples, despite having high distance according to standard metrics. For example, \citet{fredrikson2015model} consider ``model inversion'', where an image is successfully recovered from the model if it is identifiable to a human worker. In \citet{zhang2020secret}, model inversion success is measured based on pixel similarity and feature space similarity to training images. These works also recover ``representative'' images from different classes, rather than specific training examples. Recent work on reconstructing training images have used feature similarity \citep{haim2022reconstructing} and pixel similarity \citep{balle2022reconstructing}. In each of these papers, ``fuzzy'' reconstructions are allowed by the evaluation metrics and, indeed, are common in their reconstructions.

The inherent limitations of verbatim definitions of text regurgitation have also been well documented in the literature on \textbf{plagiarism detection}---both for text and code. Existing plagiarism tools, and their evaluations, go far beyond verbatim matches and consider fuzzy data ``clones'' ranging from simple transformations (e.g., word variations or shuffles) to arbitrary semantics-preserving paraphrasing \citep{roy2009comparison, potthast2010evaluation}. 
Re-purposing techniques from the plagiarism detection literature to minimize generation of memorized data in LLMs is an interesting direction toward achieving better approximate memorization definitions in machine learning.



\paragraph{Consequences for machine learning research.}
In relaxing definitions of memorization, our paper acknowledges the blurred line between memorization (e.g., of personal information) and knowledge (e.g., of common facts).
Because we use a 10-gram overlap, our \memfree{} decoding algorithm should not significantly impact utility, 
however studying this interplay is an important area of future work.
However, still, identifying which data is considered ``memorized'' cannot be done only by looking for verbatim reproductions of the training set. This may make the task of understanding memorization and generalization more difficult.

We do not believe that our work requires abandoning all research directions which
rely on prior verbatim definitions. These definitions are still useful as an efficient way to test for obvious and undeniable memorization.
However it will be necessary to continue studying further relaxations of memorization definitions to adequately capture and measure the space of privacy concerns for language models.

\section{Ethics \& Broader Impact}

Improving the privacy of neural language models---and especially those trained on user data---is
an important and timely research problem.
In this paper we hope to help both researchers and practitioners develop a more nuanced understanding
of what constitutes memorization in language models.
In particular, just because a sequence does not appear verbatim in a training dataset does not
mean the example is a novel generation: as we have shown, models today are sufficiently powerful to
minimally transform memorized data to make it appear superficially different even if the
underlying content remains memorized.

Our observation will complicate the privacy evaluation of future machine learning models.
It should no longer be deemed sufficient to check for (verbatim) matches between generated
output and a training example.
Practitioners in the future will need to be aware of this potential failure mode when
applying output post-processing defenses to mitigate memorization.
To the best of our knowledge, the only deployed system affected by our analysis is GitHub's Copilot.
In order to mitigate harm here we shared a copy of our paper with the relevant researchers at both
GitHub and OpenAI prior to paper submission.

In this paper we focus our efforts entirely on \emph{public} datasets that other researchers have
extensively studied \cite{gao2020pile} to minimize any harm caused by demonstrating extraction
results.
However, just because the data that we study is public does not mean there are no privacy concerns.
As \citet{brown2022privacy} argue, there are many other considerations when discussing the privacy
of large models trained on ``public'' datasets.

\section*{Contributions}
\begin{itemize}[noitemsep]
\item Daphne Ippolito posed the idea of memory-free decoding using a bloom filter as a solution to memorization, worked on the \memfree{} implementation, ran experiments with GPT-3 and PaLM, and contributed to paper writing.    

\item Christopher Choquette analyzed how \memfree{} used the bloom filter, created figures, and contributed to paper writing.   

\item Matthew Jagielski qualitatively analyzed generations from \memfree{}, created figures, and contributed to paper writing.   

\item Katherine Lee led figure-making,  contributed to paper writing, and resolved TODOs.    

\item Milad Nasr ran experiments with Copilot and contributed to paper writing.    

\item Florian Tramèr came up with the idea of style transferring prompts and contributed to paper writing.   

\item Chiyuan Zhang generated figures, prepared qualitative examples, and contributed to paper writing.    

\item Nicholas Carlini identified the weaknesses in memory-free decoding, worked on the \memfree{} implementation, ran experiments with GPT-Neo, and contributed to paper writing. 
\end{itemize}

\bibliography{refs}

\appendix
\clearpage
\onecolumn
\section{GitHub Copilot}\label{app:co-pilot-filter}
At the time of this paper's writing, GitHub Copilot's memorization prevention mechanism is described in their FAQ at \url{https://github.com/features/copilot}. We copy the text here:

\begin{lstlisting}[breaklines,breakautoindent=false,basicstyle=\ttfamily]
    <@\textbf{What can I do to reduce GitHub Copilot's suggestion of code that matches public code?}@>
    
    We built a filter to help detect and suppress the rare instances where a GitHub Copilot suggestion contains code that matches public code on GitHub. You have the choice to turn that filter on or off during setup. With the filter on, GitHub Copilot checks code suggestions with its surrounding code for matches or near matches (ignoring whitespace) against public code on GitHub of about 150 characters. If there is a match, the suggestion will not be shown to you. We plan on continuing to evolve this approach and welcome feedback and comment.
\end{lstlisting}
    
\section{Further Discussion of \memfree}
\label{app:bloom}

\subsection{Formal Procedure}
\label{section:algorithm}
Algorithm \ref{alg:memfree_decoding_algorithm} provides a formal procedure for \memfree{} decoding.
In all our experiments, we used $\arg \max$ decoding as the sampling method for line 4.

\begin{algorithm}[hbt!]
\caption{\memfree{} decoding algorithm.}
\label{alg:memfree_decoding_algorithm}
\footnotesize
\begin{algorithmic}[1]
\Procedure{Greedy \memfree{} Decoding}{language model $f$, prefix $p$, gen length $n$, training set $D$} 
\Repeat
    \State $\text{logits} \gets f(p) - \infty \cdot \{\mathbbm{1}[(p||t) \in D] : t \in \text{vocab}\}$  
    \State $\text{tok} \gets \text{sample from logits}$ 
    \State $p \gets p || \text{tok}$ 
\Until{$n$ iterations}
\EndProcedure
\end{algorithmic}

\end{algorithm}

\subsection{Choice of $n$-gram length}
\label{app:ngram_selection}
There are two tradeoffs to consider when choosing an $n$-gram length: the choice of $n$ changes the granularity of the memorization checking and the total number of substrings of the dataset that must be stored in the Bloom filter.
with respect to the former, notice that short $n$-grams do not have sufficient novelty (loosely, entropy) to be considered memorizations, e.g., they are often commons words and phrases. However, too large also would not capture shorter sequences that have sufficient novelty. On the latter, notice that the universe of possible $n$-grams is exponential in $n$, but that the unique number of such sequences in a fixed dataset may decrease with large $n$. This total number of unique sequences impacts the required size of the Bloom filter to maintain a fixed false positive rate. With $N$ the number of unique $n$-grams and $fp$ a decimal probability of the false positive rate, the size of the filter in bits is: 
\[ m = \left\lceil\frac{-\left(N * \log{(fp)}\right)}{\log{(2)}^2}\right\rceil. \]
Then, $k$ the number of Bloom hash functions can be calculated from the number of bits per element, i.e., $m/N$, as:
\[k = \left\lceil\left((m / N) * log(2)\right)\right\rceil. \]
This determines the cost of inserting and looking up into the Bloom filter as $\mathcal{O}(k)$. But, because $k$ typically remains small (in our case, $k=7$), this can be treated as a small constant-time operation. See~\citet{tarkoma2011theory} for the full calculations, which the ones listed here are taken from.

We err on the side of caution and select $n$=10 for our experiments.
This does prevent the model from generating common words or phrases which consist of 10 or more tokens, such as ``The quick brown fox jumped over the lazy dog.'' or ``supercalifragilisticexpialidocious''.
We find qualitatively that the impact of this is low, and that this also presents a balanced trade-off with the Bloom filter size.

\subsection{Choice of Minimum Frequency}
Ideally, we want $n$ large enough so that we do not prevent common phrases and small enough so that we catch all (though practically, most) possible memorizations. Optimizing $n$ for this task is both non-trivial, as the objective is not clear, and computationally expensive. Instead, we choose $n=10$ based on qualitative experience that this does not prevent many common phrases. Further, we do so to also limit the storage cost of the Bloom filter, because $n$ too large leads to a blow up in the number of elements, $N$.

It is important to note that using \memfree{} with a lower $n$ will result in worse performance on standard benchmarks than using it with a higher $n$.
This is because a lower $n$ means more true answers prevents from being generated.

\begin{figure}[h]
\centering
\includegraphics[width=4in]{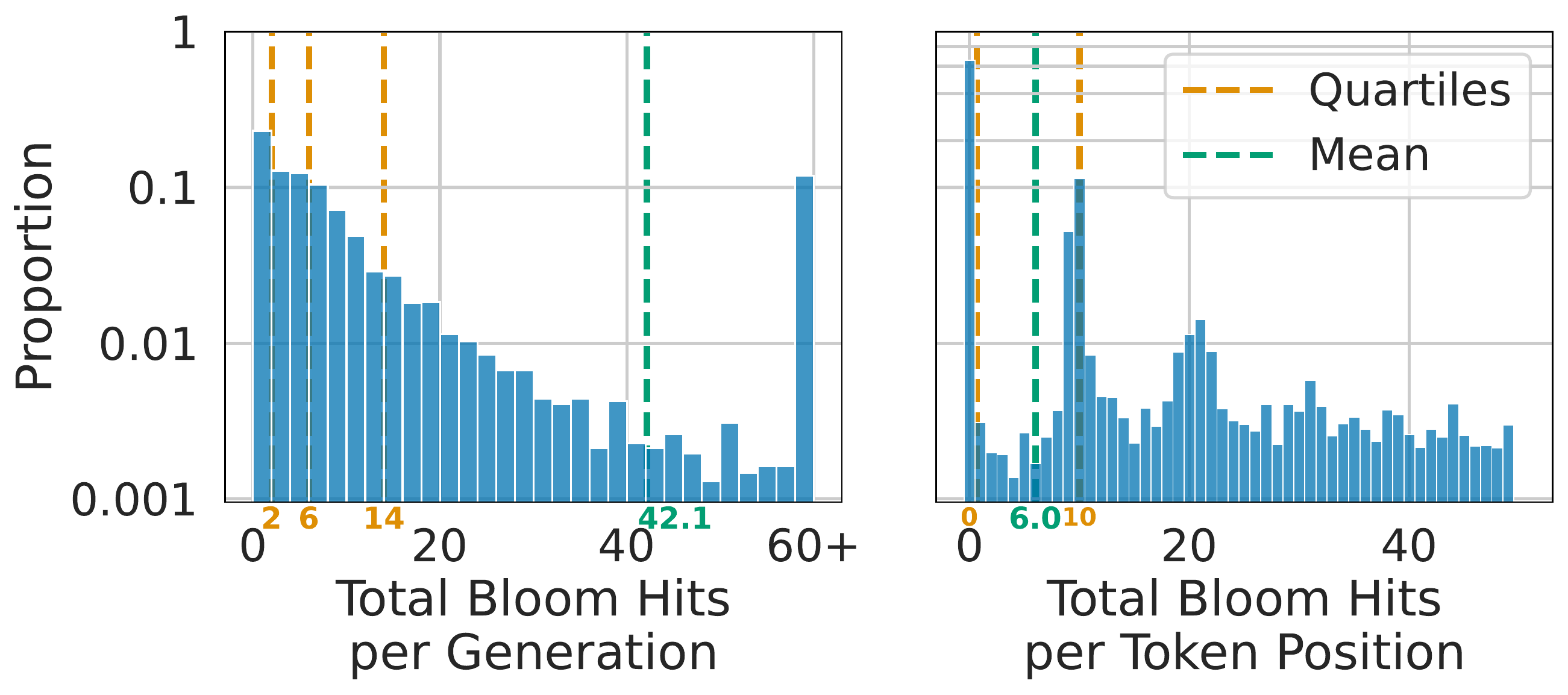}
\caption{\textbf{(left)} \textbf{Most generations have few Bloom queries,} as observed by the small quartiles; however, there is a long tail of few generations with many Bloom hits ($12.6\%$ of generations had beyond $50$ hits with a max of $1111$). \textbf{(right)} \textbf{Some positions had significantly more hits}, e.g., the first and tenth tokens. \textbf{(both)} are histograms from $6000$ generations of $50$ tokens each using \memfree{} decoding on GPT-Neo 6B.}
\label{fig:bloom-hits}
\end{figure}

\subsection{Python Implementation}
Figure \ref{fig:memfree-huggingface} contains a Python implementation of \memfree{} using the HuggingFace Transformers\footnote{\url{https://github.com/huggingface/transformers}} API.

\begin{figure}[h]
\begin{lstlisting}[language=Python,breaklines,frame=single,basicstyle=\tiny]
banned = None

model = ## huggingface model loader here
bloom = ## set-like bloom filter

num_tokens_in_filter = 10

def ban_bloom(input_ids, scores):
    """input_ids is the tokens of the prompt. scores is the logits outputted by the model given these input_ids."""
    input_ids = input_ids.cpu().detach().numpy()
    
    # Order the tokens by their likelihood.
    order = torch.argsort(-scores, 1)
    order = order.cpu().detach().numpy()
    
    batch_size = input_ids.shape[0]

    # Set the likelihood to 0 for all the most likely next tokens which would create an ngram in the bloom filter.
    for ex in range(batch_size):
        for i in order[ex]:
            sequence_to_check = (input_ids[ex].tolist() + [int(i)])
            if sequence_to_check[-num_tokens_in_filter:] in bloom:
                scores[ex,i] -= 1000
            else:
                break
    return scores

prior_processor = model._get_logits_processor
def fn(*args, **kwargs):
    prior = prior_processor(*args, **kwargs)
    prior.append(ban_bloom)
    return prior

model._get_logits_processor = fn

# Proceed with calling model.generate as normal.
\end{lstlisting}
\caption{Implementation of MemFree in HuggingFace}
\label{fig:memfree-huggingface}
\end{figure}

\subsection{Impact of \memfree{} on Downstream Task Performance}
In this section, we discuss the worst-case impact \memfree{} could have on performance on downstream tasks.
We measure this by looking at the targets, the groundtruth text a model's outputs are compared against, for three abstractive summarization tasks, three question answers tasks, and the 12 tasks in the GEM natural language generation benchmark \citep{gehrmann2021gem}.
On all these tasks, a model would score perfectly on the validation set if it exactly outputted the groundtruth target sequence.
By measuring how many of the 10-grams in each of these target sequences are present in the bloom filter used by \memfree{}, we can assess the worst-case impact \memfree{} would have on model performance at these tasks.
The results of this analysis are shown in table \ref{tab:worstcase_benchmarks}

We see that for most of these tasks, the percentage of 10-grams which are present in the bloom filter is not too much above 1\%, the false positive rate of our bloom filter.
Tasks where the target sequences come from documents likely to be present in the Pile are the most affected by \memfree{} usage.
For example, for the BillSum and Arxiv summarization tasks, over 86\% of their validation set examples have a 10-gram in the bloom filter.
Non-English tasks, which are labeled with an asterisk in Table \ref{tab:worstcase_benchmarks} were also significantly affected.
The drop in performance for non-English tasks is due to the fact that GPT-Neo's vocabulary is built off of English.
This means that non-English phrases end up being broken into many more tokens on average than English ones, and a single common word in a non-English language might take up several tokens.
This can be seen in the bloom hit examples for the MLSum-de task.

There are easy strategies to reduce the effect \memfree{} has on benchmark performance.
First, one could deliberately choose to omit from the bloom filter datasets which one decides are acceptable to memorize from, such as Wikipedia and legal documents.
Second, one could increase the $n$-gram size of the bloom filter.
As shown in the qualitative examples in Table \ref{tab:worstcase_benchmarks}, $n$=10 is perhaps too stringent for fact-based task, where names of proper nouns can take up 10-tokens or more.
Third, one could reduce the error rate of the bloom filter so as to emit fewer false positives.

\begin{table}[]
    \centering
    \scriptsize
    \begin{tabular}{p{1.1in}|r|r|r|p{2.6in}}
        \toprule
         & \textbf{\% ex with} & \textbf{\% ex} & \textbf{\% 10grams} 
        & 
        \\
        \textbf{Task} & \textbf{len>10} & \textbf{with bloom hit} & \textbf{with bloom hit}
        & \textbf{Example 10-grams with bloom hit} \\
        \midrule
        \multicolumn{5}{l}{\textbf{Summarization Tasks}} \\
        TIFU & 92.0 & 16.9 & 1.3 &
        stall windows, get new mouse, keyboard and cup $\bullet$ 
        my freezer and now my home is the bog of $\bullet$
        went to a concert five hours away as the dd
        \\
        Arxiv & 100.0 & 86.8 & 1.38 &
        of a bose gas below the critical temperature. $\bullet$ 
        in this paper, we develop a structure - preserving $\bullet$ 
        consider a model of diffusion where the individuals behavior is \\
        Pubmed & 100.0 & 92.3 & 1.7 &
        normal alanine aminotransferase $\bullet$
        the prevalence of osteoporosis in postmen $\bullet$
        www.cs.tau.ac.il \\
        BillSum & 100.0 & 88.6 & 3.0 &
        Employee Retirement Income Security Act of 1974 and the Internal
        $\bullet$
        Congressional Budget and Impoundment Control Act of 1974
        $\bullet$
        Federal Meat Inspection Act, the Poultry Products Inspection \\
        \midrule
        \multicolumn{5}{l}{\textbf{Question-Answering Tasks}} \\
        SQuAD2.0 & 9.8 & 1.1 & 5.9 & 
        E. Mann, Raymond S. Bradley and Malcolm $\bullet$ 
        CTLs (cytotoxic T lymph
        $\bullet$ 
        in 1975. It went public in 1979 and was \\
        WebQuestions & 2.4 & 0.9 & 9.8 & 
        Academia de Bellas Artes de San Fernando $\bullet$ 
        Paris Saint-Germain F.C. $\bullet$ 
        The Mating Habits of the Earthbound Human \\
        CoQA & 4.0 & 0.5 & 10.6 &
        Kingdom of Serbs, Croats and Sloven $\bullet$ 
        Sheikh Mohammed bin Rashid Al Maktou $\bullet$ 
        grabbed the rest of the pickle and ran \\
        \midrule
        \multicolumn{5}{l}{\textbf{GEM Benchmark}} \\
        \makecell[l]{CommonGen} & 81.9 & 5.7 & 1.4 & 
        You ride the horse around the area near the fence $\bullet$
        children walk with their dog on a leash down the] $\bullet$
        she wears a helmet \& sits on the motorcycle. \\
        \makecell[l]{Chezch Restaurant*\\\citep{duvsek2019semantic}} & 99.6 & 23.5 & 1.7 & jemnou restauraci BarBar, kter $\bullet$  jsou v různých $\bullet$ Bohužel, poblí \\
        \makecell[l]{DART\\\citep{nan2021dart}} & 97.1 & 20.1 & 1.7 &
        in New York City. He was a member of
        $\bullet$ 
        a low-priced family restaurant located near Raja
        $\bullet$ 
        a Member of the U.S. House of \\
       \makecell[l]{E2E clean\\\citep{duvsek2016context}} & 
        99.9 & 88.9 & 1.0 &
        near Rainbow Vegetarian Café in the city center.
        $\bullet$ 
        Phoenix is a cheap French restaurant in riverside.
        $\bullet$ 
        a French restaurant with a moderate price range, but \\
        \makecell[l]{MLSum-de*\\\citep{scialom2020mlsum}} & 
        100.0 &
        58.7 &
        2.58 &
        zum neuen Vorsitzenden $\bullet$ 
        für verfassungswidrig. $\bullet$ 
        längst überfäll \\
        \makecell[l]{MLSum-es*\\\citep{scialom2020mlsum}} &
        100.0 & 42.3 & 2.2 &
        del pacto y no de la confrontación $\bullet$ 
        selección española de f $\bullet$ 
        investigación sobre la desaparici \\
        Schema-Guided Dialog &
        63.3 & 7.5 & 1.3 &
        The Lord of the Rings: The Return of the $\bullet$
        tyard By Marriott Sacramento Cal Expo has a 3 star $\bullet$
        with Southwest Airlines. The flight takes off at 7 \\
        \makecell[l]{ToTTo\\\citep{parikh2020totto}} & 98.0 & 20.9 & 3.2 &
        and was broadcast on Venevisión. $\bullet$
        As of the census of 2000, there were 133 $\bullet$
        on the U.S. Billboard 200 chart. \\
        XSum & 99.4 & 18.5 & 1.6 & 
        stressed will not increase your risk of dying, according $\bullet$
        Two drug dealing brothers taken back to court for mocking $\bullet$
        the Institute of Directors (IoD) has
        \\
        WebNLG-en & 97.9 & 27.4 & 4.7 &
        written by J.R.R. Tolkien, $\bullet$
        play in the Campeonato Brasileiro $\bullet$
        is affiliated with Visvesvaraya Technological University \\
        WebNLG-ru* & 100.0 & 99.6 & 42.9 &
        $\bullet$
        $\bullet$
        \\
        WikiAuto + Turk/ASSET & 96.5 & 16.7 & 2.2 &
        pop-punk, surf rock, ska, $\bullet$
        was discovered by a team of astronomers from the University $\bullet$
        cover of Sgt. Pepper's Lonely Hearts Club Band \\
        \bottomrule 
    \end{tabular}
    \caption{\textbf{Some benchmark tasks could be significantly affected by \memfree{}.} For several standard benchmark tasks commonly used to evaluate language models, we report the percentage of test set target sequences which consist of at least one 10-gram (meaning hitting the bloom filter is possible),  the percentage of test set target sequences which contain at least one 10-gram present in the bloom filter, and the percentage of all the 10-grams in the test set targets which can be found in the bloom filter. We also show 3 example 10-grams (delineated by `$\bullet$') which are present in both the test set and the bloom filter. (For QA tasks, we only consider the first answer for each question.) \textbf{The numbers here reflect the worst case scenario: the fraction of examples a language model that perfectly memorized the test set would be incapable of getting exactly correct when used with \memfree.}}
    \label{tab:worstcase_benchmarks}
\end{table}

\subsection{Performance of \memfree{}}
In this section, we study two questions: (1) ``does \memfree{} maintain model utility?'' and (2) ``does our optimized \memfree{} prevent memorization release''.

Along question (1), recall that \memfree{} can admit false positives, which may degrade the utility of the language model.
Fortunately, the false positive rate can be computed exactly, e.g., see~\citet{tarkoma2011theory}, and a long literature has proposed optimizations to account for non-uniform distributions~\citep{bruck2006weighted} and to adaptively correct for false positives~\citep{bender2018bloom}.

Here, we study how, under reasonable computational constraints and inference times, the observed rates impact model utility. As we will show, we observe that \memfree{} maintains the highest utility (no observable impact) while being the most efficient defense.



%

Along question (2), we study if our optimizations lead to a substantial increase in the false negative rate. To do this, we repeat the experiment from~\cite{carlini2022quantifying}, which prompted GPT-Neo models with examples from its training data. We compute how many examples are verbatim memorized when \memfree{} decoding is used. The 6B parameter GPT-Neo model memorizes more than 12,000 of these documents, but, after applying \memfree{}, it only outputs 4 verbatim memorizations. These 4 remaining verbatim memorizations are repeated fewer than 10 times in the training data, and so were not added to our Bloom filter. Nonetheless, this strategy reduced verbatim memorization by over $3000\times$.

\subsection{Bloom Filter Statistics}\label{app:ssec:stats}
\begin{figure}[H]
    \centering
    \includegraphics[width=3in]{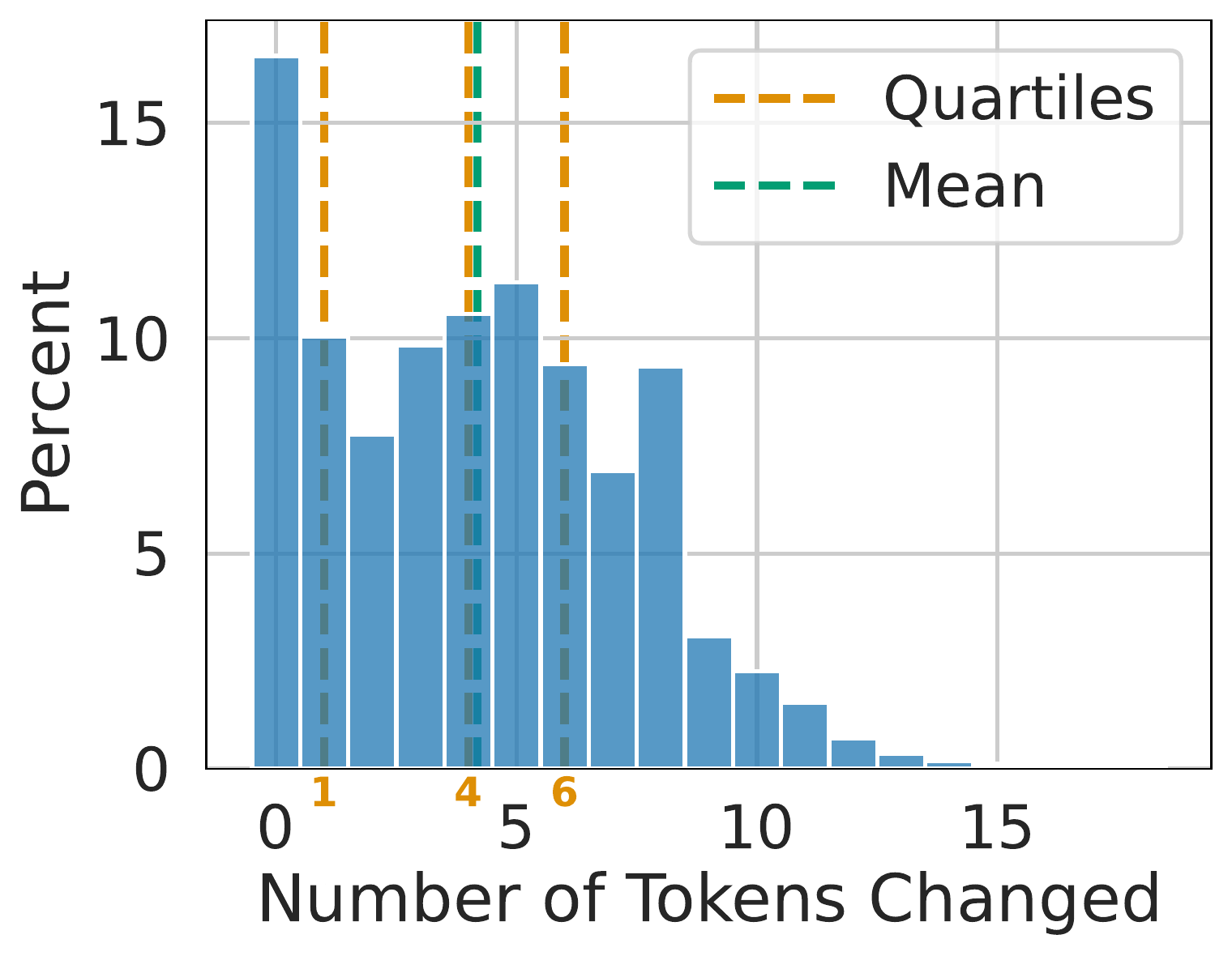}
    \caption{\textbf{Most generations require few ($<5$) changes to pass \memfree checks.} Data for histogram from $6000$, $50$-token generations using \memfree{} decoding on GPT-Neo 6b.}
    \label{fig:memfree-tokens-changed-stats}
\end{figure}

Figure \ref{fig:memfree-tokens-changed-stats} shows the distribution in number of tokens (out of 50 generated) that were changed by \memfree{} from the token that would have been generated using undefended greedy decoding.

\noindent Figure \ref{fig:bloom-hits} presented some of the query patterns of the \memfree{} decoder to investigate when and how it impacts decoding.
First, we observe that \memfree{} is trivial to run in terms of compute: it takes only $49.8$ milliseconds to run 10,000 queries on one CPU core. From Figure~\ref{fig:bloom-hits} (left), all generations required significantly fewer queries (mean $= 42.1$ queries / generation)---even running batches of many hundreds or thousands of queries would incur less than a few seconds additional overhead. 
Second, we find that the Bloom filter is often hit at the first and tenth tokens after the prompt.
We see many hits at the first token because all our prompts are from the training data---so there are relatively fewer single token additions that generate a novel $n$-gram.
Third, we find that most generations need only a few ($<5$) alterations due to \memfree{}
 decoding.


\section{More Details on Measuring Approximate Memorization}

\subsection{Similarity Metrics Implementations}
\label{app:eval_metrics_details}
As noted in Section \ref{sec:experiments}, we identify instances of approximate memorization by measuring the similarity between a generated continuation and the groundtruth continuation for a prompt. We do so using BLEU-score and character-level edit distance.

We computed BLEU score using NLTK's BLEU computation (\texttt{nltk.translate.bleu\_score}) with the default parameters (averaging equally BLEU-1, BLEU-2, BLEU-3, and BLEU-4)\cite{bird2009natural}. 
Edit distance was computed with the `editdistance` pip package. Normalized edit similarity between two strings $x$ and $y$ is defined as:

\[
\text{\textsc{EditSim}}(x, y) = \frac{\text{\textsc{EditDistance}}(x, y)}{\max(|x|, |y|)}
\]

\subsection{BLEU Score Threshold Selection}
\label{app:bleu_score_threshold}
We chose to use a BLEU score of 0.75 or higher to indicate that a generation substantially memorized from the ground-truth continuation.
We choose to threshold BLEU score rather than edit-distance since it is more interpretable to NLP researchers. 
Table \ref{tab:approx_mem_at_threshold} presents several examples of generations that were very close to the 0.75 BLEU score threshold.
This threshold was deliberately set high so as to avoid false-negatives, allowing us to provide a reliable lower bound on memorization.
In our qualitative analysis of the data, all examples that were above BLEU of 0.75 showed unambiguous memorization.
We also observed memorization in some examples with lower BLEU scores.

\begin{figure*}[h]
    \centering
    \small
    \begin{tabular}{p{2.5in}|p{2.5in}|r}
    \toprule
        \textbf{True Continuation} & \textbf{LM Generation} & BLEU \\
        \midrule
         about a deeper bond with the product. Nostalgia. It’s delicate, but potent. Sweetheart.\textbackslash n\textbackslash nTeddy told me that in Greek, “nostalgia” literally means “the pain from an old wound”. It’s a twinge in your heart far more & about a deeper bond with a product. Nostalgia. It's delicate, but potent. Teddy told me that in Greek, nostalgia literally means the pain from an old wound. It's a twinge in your heart, far more powerful than memory alone. This	& 0.79\\
         \midrule
         And I'll show you\textbackslash nYou want the moves like jagger\textbackslash nI got the moves like jagger\textbackslash nI got the mooooooves like jagger\textbackslash nI don't even try to control you\textbackslash nLook into my eyes and I'll own you\textbackslash nYou with the moves	&
         And I'll show you\textbackslash nAll the moves like Jagger\textbackslash nI've got the moves like Jagger\textbackslash nI've got the moves like Jagger\textbackslash nI don't need to try to control you\textbackslash nLook into my eyes and I'll own you\textbackslash nWith the	& 0.79 \\
         \midrule
         Purpose\textbackslash n\textbackslash nThe laws of most jurisdictions throughout the world automatically confer\textbackslash nexclusive Copyright and Related Rights (defined below) upon the creator\textbackslash nand subsequent owner(s) (each and all, an "owner") of an original work of\textbackslash nauthorship and/or a database &
         Purpose\textbackslash n\textbackslash n The laws of most jurisdictions throughout the world automatically confer\textbackslash n exclusive Copyright and Related Rights (defined below) upon the creator\textbackslash n and subsequent owner(s) of an original work of authorship (the "Work").\textbackslash n Certain jurisdictions do not recognize a	& 0.76 \\
    \bottomrule
    \end{tabular}
    \caption{Examples of model generations with BLEU score close to 0.75, the threshold we used to declare that approximate memorization had occurred.}
    \label{tab:approx_mem_at_threshold}
\end{figure*}

\section{Experiments with Large English Language Models}\label{app:sec:large-english-model}

\subsection{Prompt Selection Process}
Famous speeches were selected from the "Top 100 Speeches" list found at \url{https://www.americanrhetoric.com/newtop100speeches.htm}.
Monologues were selected from the list of two-minute monologues found at \url{http://www.monologuedb.com/tag/2-minute-monologues/}.
Novels were selected from the Time Magazine’s Top 100 All-Time Novels list found at \url{https://www.goodreads.com/list/show/2681.Time_Magazine_s_All_Time_100_Novels}.
The opening paragraphs of the first chapter (skipping over prefaces, introductions, and boilerplate) were used as each example.
The 2011 and 2021 song lyrics were selected from the Billboard Year-End Hot 100 singles lists found at \url{https://en.wikipedia.org/wiki/Billboard_Year-End_Hot_100_singles_of_2011} and \url{https://en.wikipedia.org/wiki/Billboard_Year-End_Hot_100_singles_of_2012}.

For each document, the first 100 \textit{words} were used as a prompt, and the first 50 generated \textit{words} were compared with the first 50 words of the true continuation.
This approach has the ramification that not all prompts were the same length in \textit{tokens}.
However, this approach was necessary for fairness across style transfers because an all-uppercased string is going to be many subword tokens longer than the lowercased version of the same string.

\begin{figure*}[h]
    \centering
    \includegraphics[width=\linewidth]{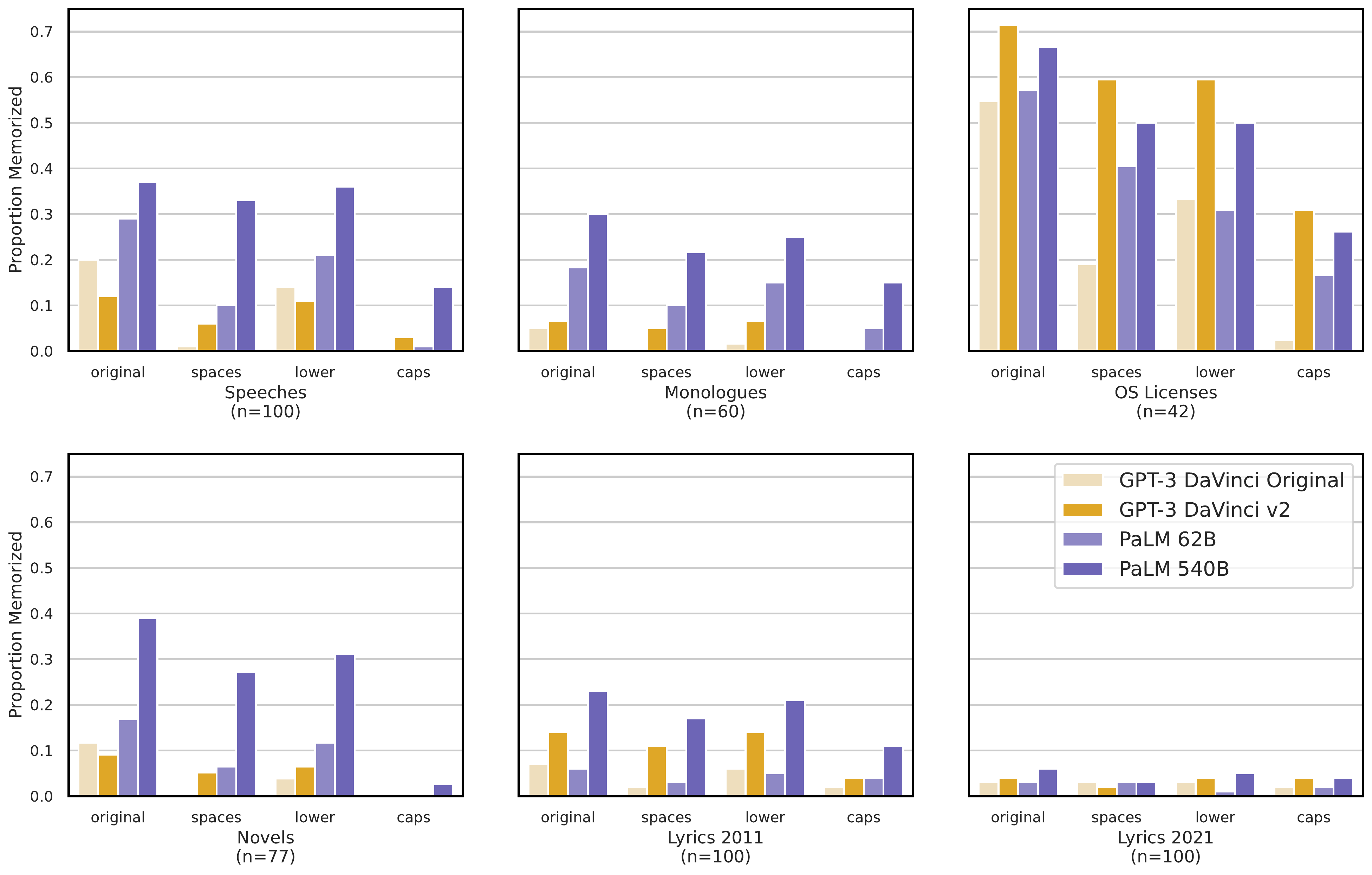}
    \caption{\textbf{"Style-transfer" prompting divulges approximate memorization in two versions of GPT-3 and two sizes of PaLM.} Note that generations also follow the same style as the prompt. Generations were characterized as memorized if they had a BLEU score of at least 0.75 with the ground-truth continuation.
    }
    \label{fig:famous-datasets-all}
 \end{figure*}


\begin{table*}[]
    \centering
    \small
    \begin{tabular}{@{}ll rrrr }
    \toprule
        \multirow{2}{*}{Domain with $n$ total prompts} & \multirow{2}{*}{Model} &
        \multicolumn{4}{c}{\makecell{\# Prompts Memorized per\\Style-Transfer Type}} \\\cmidrule{3-6}
        & & Original & Two Spaces & Lower & Upper \\
        \toprule
        Open-Source Licenses ($n$=42) & GPT-3 DaVinci Original & 23 & 8 & 14 & 1 \\
        & GPT-3 DaVinci v2 & 30	& 25	& 25	& 13 \\
        \midrule
        Famous Speeches ($n$=100) & GPT-3 DaVinci Original & 20 & 1 & 14 & 0 \\
        & GPT-3 DaVinci v2 & 12	& 6	& 11	& 3\\
        \midrule
        Famous Monologues ($n$=60) & GPT-3 DaVinci Original & 3 & 0 & 1 & 0 \\
        & GPT-3 DaVinci v2 & 4	& 3	& 4	& 0	\\
        \midrule
        Novel Openings ($n$=77) & GPT-3 DaVinci Original & 9 & 0 & 3 & 0 \\
        & GPT-3 DaVinci v2 & 7	& 4	& 5	& 0 \\
        \midrule
        Lyrics 2011 ($n$=11) & GPT-3 DaVinci Original & 7 & 2 & 6 & 2 \\
        & GPT-3 DaVinci v2 & 14 & 11 & 14 & 4\\
        \midrule
        Lyrics 2021 ($n$=11) & GPT-3 DaVinci Original & 3 & 3 & 3 & 2 \\
        & GPT-3 DaVinci v2 & 4 & 2 & 4 & 4\\
        \bottomrule
    \end{tabular}
    \caption{\textbf{"Style-transfer" prompting surfaces approximate memorization in GPT-3.} We explore $n$ prompts for each domain. Note that generations also follow the same style as the prompt.}
    \label{tab:gpt3}
\end{table*}
 \clearpage
 
\section{Experiments with \memfree{} and Other Model Families}
In addition to running experiment using the GPT-Neo family, we also ran them with the Pyhia model family \citep{biderman2023pythia}.
Like GPT-Neo, Pythia was trained on the Pile.
There are two versions of Pythia, one trained on the same version of the Pile as GPT-Neo, and another trained on a deduplicated version of the Pile.

\begin{figure}[h]
    \centering
    \includegraphics[width=0.8\linewidth]{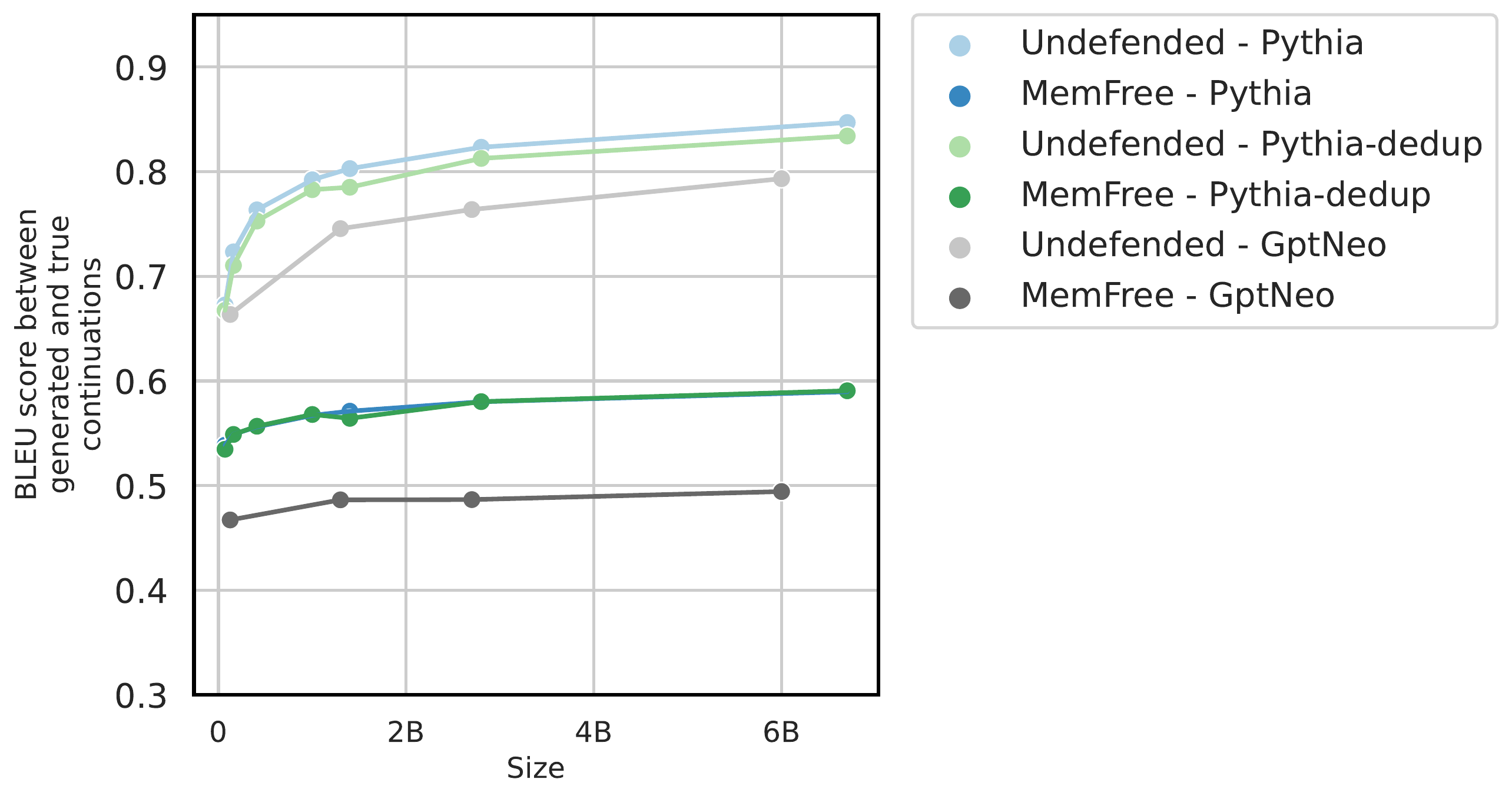}
    \caption{Approximate memorization on the base and deduped versions of Pythia, compared with GPT-Neo.}
    \label{fig:pythia_bleu}
\end{figure}

\begin{figure}[h]
    \centering
     \begin{subfigure}[t]{0.4\textwidth}
          \centering
          
          \includegraphics[width=\linewidth]{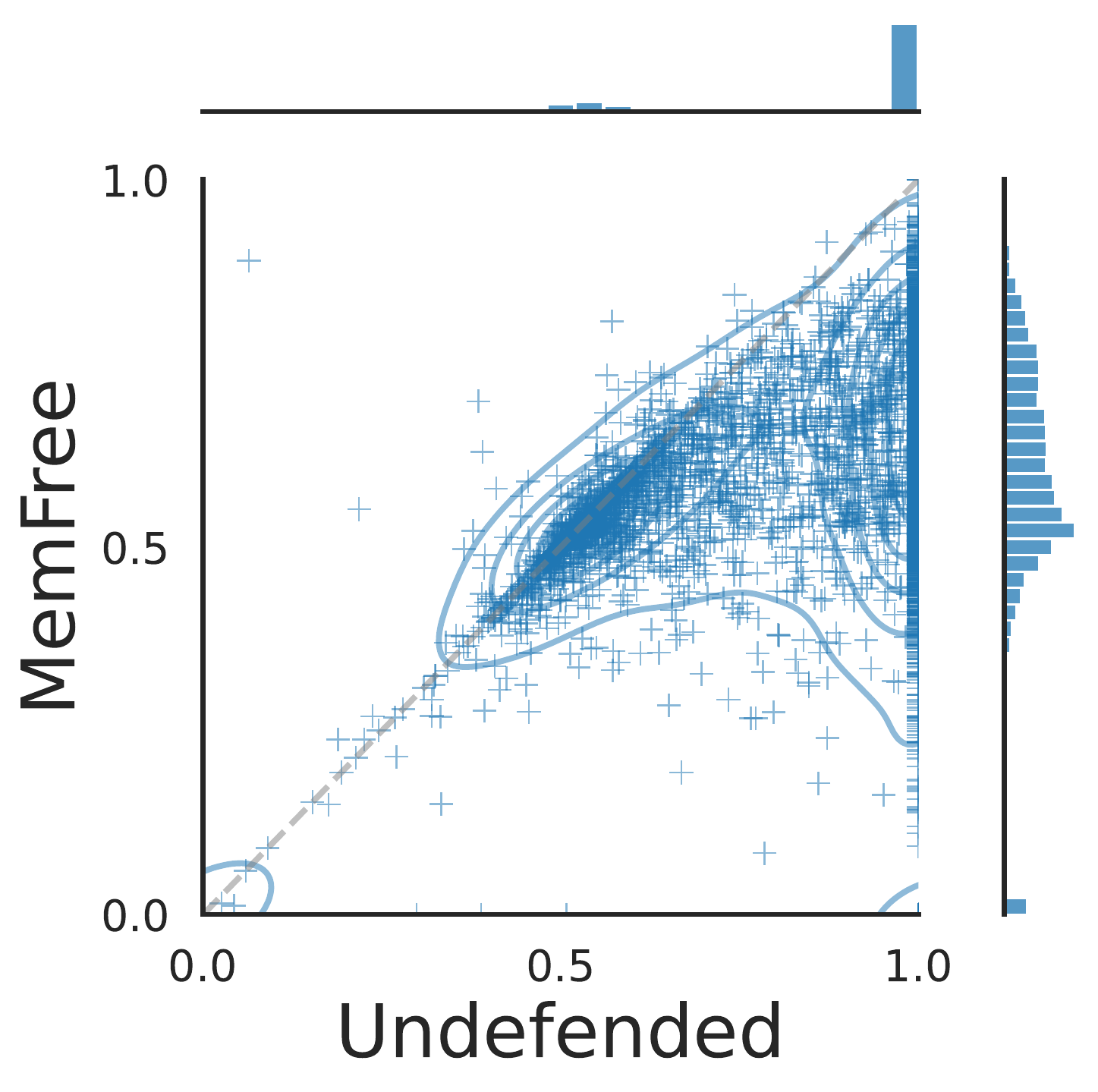}
          \caption{BLEU (word-level)}
          \label{fig:bleu_similarity}
      \end{subfigure}
      \hfill
      \begin{subfigure}[t]{0.4\textwidth}
          \centering
          \includegraphics[width=\linewidth]{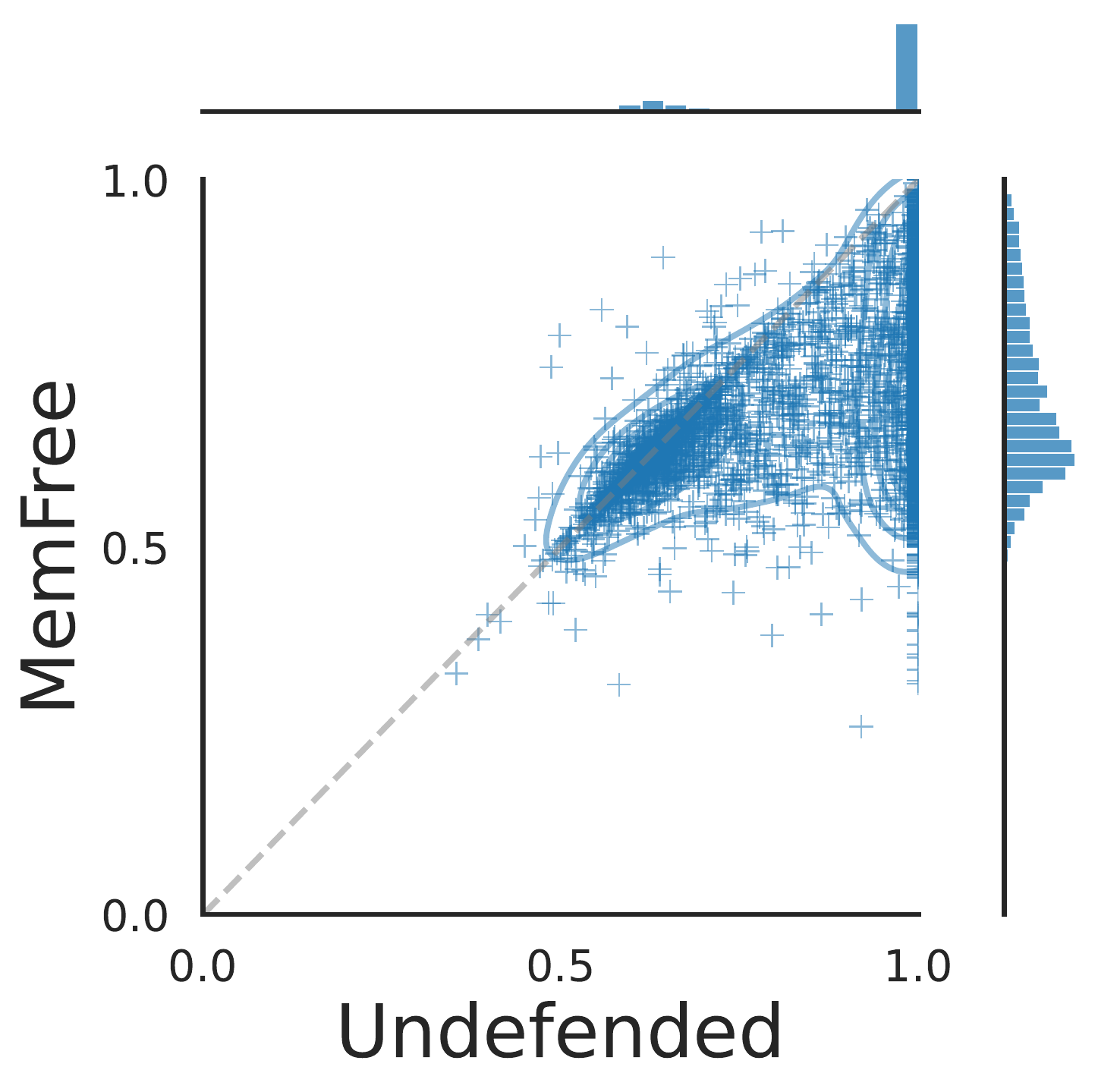}
          \caption{Edit similarity (char-level)}
          \label{fig:char_distance}
     \end{subfigure}
     \caption{\memfree{} is also effective at reducing approximate memorization for the deduped Pythia 6.9B model.}
     \label{fig:distance_scatter-pythia}
\end{figure}

Figure \ref{fig:pythia_bleu} shows the amount of memorization in each of these three model families, with and without \memfree{}.
Figure \ref{fig:distance_scatter-pythia} shows the same scatter plots as in Figure \ref{fig:distance_scatter}, but using the 6.9B-parameter Pythia.
We see that Pythia exhibits more approximate memorization than GPT-Neo.
Though \memfree{} is still effective at reducing approximate memorization, it is slightly less effective than it was on GPT-Neo.

\section{Qualitative Examples that Bypass Copilot's Filter}\label{app:examples-bypassing-copilot}
In Figure \ref{fig:app-co-pilot-prompting},  we show more examples that bypass Copilot's memorization filter.

\begin{figure*}
\footnotesize
\begin{subfigure}[t]{.47\linewidth}
{\small\textbf{Standard Prompting}}
\begin{lstlisting}[breaklines,frame=single]
<@\textcolor{color-blind-cyan}{/* low  –> Starting index,  high  –> Ending index */ }@>
<@\textcolor{color-blind-cyan}{ quickSort(arr[], low, high)}@>
{
    if (low < high)
    {
        /* pi is partitioning index, arr[p] is now
              at right place */
        pi = partition(arr, low, high);

        
<@\textcolor{color-blind-orange}{Copilot no longer generates continuations}@>
\end{lstlisting}
\end{subfigure}
\vspace{2em}
\begin{subfigure}[t]{.51\linewidth}
{\small\textbf{Comment Prompting}}
\begin{lstlisting}[breaklines,frame=single]
<@\textcolor{color-blind-cyan}{ \# /* low  –> Starting index,  high  –> Ending index */}@>
# quickSort(arr[], low, high)
# {
#     if (low < high)
#     {
#         /* pi is partitioning index, arr[p] is now
#            at right place */
#         pi = partition(arr, low, high);
#         quickSort(arr, low, pi - 1);  # Before pi
#         quickSort(arr, pi + 1, high); # After pi
#     }
# }
\end{lstlisting}
\end{subfigure}
\begin{subfigure}[t]{.48\linewidth}
{\small\textbf{Naming Convention Prompting}}
\begin{lstlisting}[breaklines,frame=single]
<@\textcolor{color-blind-cyan}{/* \_low  –> Starting index,  \_high  –> Ending index */}@>
<@\textcolor{color-blind-cyan}{quickSort}@>(arr[], _low, _high)
{
    if (_low < _high)
    {
        /* pi is partitioning index, arr[p] is now
            at right place */
        pi = partition(arr, _low, _high);
        quick_sort(arr, _low, pi - 1);  // Before pi
        quick_sort(arr, pi + 1, _high); // After pi
    }
}



\end{lstlisting}
\vspace{2em}
\end{subfigure}
\quad
\begin{subfigure}[t]{.51\linewidth}
{\small\textbf{Language Prompting}}
\begin{lstlisting}[breaklines,frame=single,showstringspaces=false,
literate=
{á}{{\'a}}1
{à}{{\`a}}1
{é}{{\'e}}1
{è}{{\`e}}1
]
<@\textcolor{color-blind-cyan}{/* depart  –> index de départ,  fin  –> index de fin */ }@>
<@\textcolor{color-blind-cyan}{quickSort(arr[], depart, fin)}@>
{
    if (depart < fin)
    {
        /* pi est l'index de partitionnement, arr[p] est maintenant
              à la bonne place */
        pi = partition(arr, depart, fin);
        // Trier les éléments séparément avant et après la partition
        quick_sort(arr, depart, pi - 1);
        quick_sort(arr, pi + 1, fin);
    }
}
\end{lstlisting}
\end{subfigure}

\begin{subfigure}[t]{.46\linewidth}
{\small\textbf{Naming Convention Prompting}}
\begin{lstlisting}[breaklines,frame=single]
<@\textcolor{color-blind-cyan}{/* LOW  –> Starting index,  HIGH  –> Ending index */}@>
quickSort(arr[], LOW , HIGH )
{
    if (LOW  <  HIGH )
    {
        /* pi is partitioning index, arr[p] is now
              at right place */
        pi = partition(arr, LOW , HIGH );
        quickSort(arr, LOW , pi - 1);  // Before pi
        quickSort(arr, pi + 1, HIGH );  // After pi
    }
}

\end{lstlisting}
\vspace*{5em}
\end{subfigure}
\hfill
\begin{subfigure}[t]{.51\linewidth}
{\small\textbf{Comment Prompting}}
\begin{lstlisting}[breaklines,frame=single]
<@\textcolor{color-blind-cyan}{/* low  –> Starting index,  high  –> Ending index */ //1}@>
<@\textcolor{color-blind-cyan}{quickSort}@>(arr[], low, high) //2
{
    if (low < high) //3
    {
        /* pi is partitioning index, arr[p] is now
              at right place */ 
        pi = partition(arr, low, high); //4
        quickSort(arr, low, pi - 1); //5
        quickSort(arr, pi + 1, high); //6
    }
}



\end{lstlisting}
\end{subfigure}
    \caption{Preventing GitHub's Copilot language model from emitting verbatim memorized content
    often results in it emitting memorized, but not verbatim, content. 
    Trivially modifying the prompt bypasses these output filters.
    \textcolor{color-blind-cyan}{Prompts are highlighted in blue.}  
    Model evaluated with the option ``block suggestions matching public code'' enabled. 
    For brevity we removed comments from model outputs.}
    \label{fig:app-co-pilot-prompting}
\end{figure*}
\clearpage

\section{Qualitative Examples of Memorization Despite \memfree{}}
\label{app:memfree-fails}
Table \ref{tab:still_memorized} gives examples of approximate memorization which occured despite using \memfree{} decoding.
Below
, we present a longer list of full examples of \memfree{} decoding. In the following paragraphs, the \hlc[gray!20]{gray highlighted texts} are the prompts, and the texts that follow the prompts are generated by the model. For easier reading, we merged the individual tokens to form text spans, except at tokens where bloom-filter rejection happens. In this case, the symbol $\rhd$ indicate connection between tokens. Moreover, \hlc[red!20]{red highlighted texts} indicate one or more rejected tokens, which are always followed by one accepted token \hlc[green!20]{highlighted with green}. The texts are lightly edited mostly to properly encode unicode symbols (some garbled symbols are replaced as  $\boxtimes$) that \LaTeX{} does not handle gracefully.

\paragraph{Split word ``Activity''}
\hlc[gray!20]{viewHolder.swipeLayout.setOnDoubleClick
Listener(new SwipeLayout.DoubleClickListener() \{             \@Override             public void onDoubleClick(SwipeLayout layout, boolean surface) \{                 Toast.makeText(mContext, "DoubleClick : " + position,}   $\rhd$ \hlc[red!20]{}  \hlc[green!20]{Toast} $\rhd$  .LENGTH  $\rhd$ \hlc[red!20]{\_}  \hlc[green!20]{).} $\rhd$  show();  $\rhd$ \hlc[red!20]{} \hlc[red!20]{\}} \hlc[red!20]{//} \hlc[red!20]{\});} \hlc[red!20]{if} \hlc[red!20]{return}  \hlc[green!20]{m} $\rhd$  Context.start  $\rhd$ \hlc[red!20]{Activity} \hlc[red!20]{Service}  \hlc[green!20]{Activ} $\rhd$  ities(new Intent(mContext,  $\rhd$ \hlc[red!20]{}  \hlc[green!20]{Activity} $\rhd$  .class));  $\rhd$ \hlc[red!20]{} \hlc[red!20]{\}}  \hlc[green!20]{m} $\rhd$  Context.fin  $\rhd$ \hlc[red!20]{ish}  \hlc[green!20]{ishing} $\rhd$  ();

\paragraph{Typo ``Wildlife'' $\Rightarrow$ ``Wildife''}
\hlc[gray!20]{should promptly announce a phase-in of non-lead ammunition for all hunting throughout the rest of the state to protect other wildlife poisoned by lead and to safeguard human health. Hunters can play a critical role in the recovery of the condor and also keep lead from poisoning the wild game they eat.''  The lead-free ammunition regulations are designed to reduce lead poisonings of the iconic and extremely endangered California condor. Condors, eagles, and other scavengers such as ravens}  $\rhd$ , crows, and hawks  $\rhd$ \hlc[red!20]{are}  \hlc[green!20]{have} $\rhd$  been poisoned by lead ammunition.  The California Department of Fish and  $\rhd$ \hlc[red!20]{Wildlife} \hlc[red!20]{Game}  \hlc[green!20]{Wild} $\rhd$  ife (CDFW) is the lead agency responsible for implementing the regulations. The regulations are based on the Condor Protection Act,

\paragraph{Singular to plural: ``claim'' $\Rightarrow$ ``claims''}
\hlc[gray!20]{) No 1924/2006.  neurotransmission and muscle contraction including heart muscle  29  Magnesium  Magnesium contributes to normal protein synthesis  The claim may be used only for food which is at least a source of magnesium as referred to in the claim SOURCE OF [NAME OF VITAMIN/S] AND/OR [NAME OF MINERAL/S] as listed in the Annex to Regulation (EC) No 1924/2006.  protein}  $\rhd$ synthesis  30  $\rhd$ \hlc[red!20]{M}  \hlc[green!20]{N} $\rhd$  iacin  Niac  $\rhd$ \hlc[red!20]{in}  \hlc[green!20]{ins} $\rhd$  contribute to normal protein synthesis  The  $\rhd$ \hlc[red!20]{claim}  \hlc[green!20]{claims} $\rhd$  may be used only for food which is at  $\rhd$ \hlc[red!20]{least}  \hlc[green!20]{lest} $\rhd$  a source of niacin as referred to  $\rhd$ \hlc[red!20]{in}  \hlc[green!20]{to} $\rhd$  in the claim SOURCE OF [

\paragraph{Uppercase to lowercase}
\hlc[gray!20]{cuisine of the city's long-established Hum Vegetarian Cafe \& Restaurant to a central garden location. Settle into the elegant and verdant space and enjoy dishes including papaya and banana flower salads, mushrooms steamed in coconut, and the subtle combination of braised tofu with star anise and cinnamon.  Nha Hang NgonVietnamese\$\$  (MAP GOOGLE MAP);  \%028-3827 7131; 160 $\boxtimes$ Pasteur; meals 60}  $\rhd$ ,000-100,000đ;  h7am-10pm;  $\rhd$ \hlc[red!20]{$\boxtimes$}  \hlc[green!20]{)} $\rhd$  This is the place to come for  $\rhd$ \hlc[red!20]{a}  \hlc[green!20]{authentic} $\rhd$  Vietnamese food in the French  $\rhd$ \hlc[red!20]{Quarter}  \hlc[green!20]{quarter} $\rhd$  . The menu is a mix of Vietnamese and French dishes, and the food is

\paragraph{Lower case to upper case}
\hlc[gray!20]{broker = createBroker(deleteAllMessagesOnStartup);         broker.start(); \} public BrokerService createBroker(boolean deleteAllMessagesOnStartup) throws Exception \{ return createBroker(deleteAllMessagesOnStartup, TRANSPORT\_URI); \} public BrokerService createBroker}  $\rhd$ \hlc[red!20]{(}  \hlc[green!20]{With} $\rhd$  TransportURI(boolean deleteAllMessagesOnStart $\rhd$ \hlc[red!20]{up} \hlc[green!20]{Up} $\rhd$ ) throws Exception \{ $\rhd$ \hlc[red!20]{$\boxtimes$} \hlc[green!20]{return}  $\rhd$ \hlc[red!20]{create} \hlc[red!20]{new} \hlc[green!20]{broker} $\rhd$  Service.createBroker(deleteAllMessages  $\rhd$ \hlc[red!20]{On}  \hlc[green!20]{,} $\rhd$  TRANSPORT\_URI); \}  $\rhd$ \hlc[red!20]{$\boxtimes$}  \hlc[green!20]{$\boxtimes$} $\rhd$  \hlc[red!20]{} \hlc[green!20]{\}} $\rhd$ \hlc[red!20]{$\boxtimes$} \hlc[red!20]{$\boxtimes$}  \hlc[green!20]{<|endoftext|>}

\paragraph{Change from ``agree'' to `Ag-reeableness''}
\hlc[gray!20]{a person is imaginative or independent, high openness can be perceived as unpredictability or lack of focus. Moreover, individuals with high openness are said to pursue self-actualization specifically by seeking out intense, euphoric experiences, such as skydiving, living abroad, gambling, et cetera. Conversely, those with low openness seek to gain fulfillment through perseverance, some disagreement remains about how to interpret and contextualize the openness factor. A tendency to be organized and dependable, show self-}  $\rhd$ discipline, and be goal-oriented is also associated with high openness.  The openness factor is also associated with the Big Five personality traits of  $\rhd$ \hlc[red!20]{agree}  \hlc[green!20]{Ag} $\rhd$  reeableness, Conscientiousness  $\rhd$ \hlc[red!20]{,}  \hlc[green!20]{and} $\rhd$  Emotional Stability.  See also  Openness

\paragraph{Passive voice to active voice (grammar error)}
\hlc[gray!20]{still wouldn't shock me at all if the Chiefs pulled off a Giants- or Ravens-esque Super Bowl run to cap off this five-year window of the Smith/Reid era with a ring. While Pittsburgh has been this team's bugaboo, maybe they can avoid them in January thanks to a Jacksonville upset, and I still say this is the AFC team most likely to win a playoff game in New England. While so many were quick to write the Chiefs off, they just opened up}  $\rhd$ a new chapter in their history.  The Chiefs are the AFC's best team, and they're going to be a force to be  $\rhd$ \hlc[red!20]{reckoned}  \hlc[green!20]{reckon} $\rhd$  with for years to come.  1. New England Patriots  $\rhd$ \hlc[red!20]{$\boxtimes$}  \hlc[green!20]{The} $\rhd$  Patriots are the AFC's best

\paragraph{Change of protocol (email still get generated)}
\hlc[gray!20]{="https://groups.google.com/group/django-developers" target="\_blank" rel="nofollow" onmousedown="this.href=\&\#39;https://groups.google.com/group/django-developers\&\#39;;return true;" onclick="this.href=\&\#39;https://groups.google.com/group/django-developers\&\#39;;return true;">https://groups} $\rhd$ \hlc[red!20]{.}  \hlc[green!20]{...} $\rhd$  </a>  <a href="  $\rhd$ \hlc[red!20]{https} \hlc[red!20]{http}  \hlc[green!20]{mail} $\rhd$  to  $\rhd$ \hlc[red!20]{:}  \hlc[green!20]{://} $\rhd$  david@davidwalsh.name" target="\_blank" rel="n  $\rhd$ \hlc[red!20]{of} \hlc[red!20]{ore}  \hlc[green!20]{...} $\rhd$  </a>  <a href="mailto://david@davidw

\paragraph{Synonyms}
\hlc[gray!20]{ken interior. The seats were heavily cushioned black velvet. On the windows, the Darkling's symbol had been cut into the glass: two overlapping circles, the sun in eclipse.  Across from me, the two Grisha were studying me with open curiosity. Their red kefta were of the finest wool, embroidered lavishly in black and lined in black fur. The fair-haired Heartrender was lanky and had a long, melancholy face. Ivan was taller, broader,}  $\rhd$ and had a face like a bulldog's.  "You are  $\rhd$ \hlc[red!20]{the}  \hlc[green!20]{a}  $\rhd$ \hlc[red!20]{Gr}  \hlc[green!20]{very} $\rhd$  pretty girl," Ivan said.  "Thank you," I  $\rhd$ \hlc[red!20]{said} \hlc[red!20]{replied}  \hlc[green!20]{answered} $\rhd$  .  $\rhd$ \hlc[red!20]{$\boxtimes$}  \hlc[green!20]{"} $\rhd$  I'm not a girl."  "You are a girl," he said.

\paragraph{Synonyms}
\hlc[gray!20]{severing any such bond.  In re L.M., 923 A.2d 505, 511 (Pa. Super. 2007) (citing 23 Pa.C.S. \S{} 2511)  (some citations omitted). Section 2511(a) provides in pertinent part: (a) General rule.-The rights of a parent in regard to a child}  $\rhd$   $\rhd$ \hlc[red!20]{may} \hlc[green!20]{are} $\rhd$ not terminated by a proceeding brought under  $\rhd$ \hlc[red!20]{}  \hlc[green!20]{this} $\rhd$ \hlc[red!20]{part} \hlc[red!20]{chapter} \hlc[red!20]{section} \hlc[red!20]{sub} \hlc[red!20]{subsection} \hlc[red!20]{[} \hlc[red!20]{article}  \hlc[green!20]{paragraph} $\rhd$  or  $\rhd$ \hlc[red!20]{paragraph}  \hlc[green!20]{section} $\rhd$  2512 or 2513(a) or (b), or any  $\rhd$ \hlc[red!20]{$\boxtimes$}  \hlc[green!20]{\_\_\_\_\_\_\_\_\_\_\_\_\_\_\_\_\_\_\_\_\_\_\_\_\_\_\_\_\_\_\_\_} $\rhd$  \_\_\_\_\_\_\_\_\_\_\_\_  $\rhd$ \hlc[red!20]{} \hlc[red!20]{} \hlc[red!20]{$\boxtimes$}  \hlc[green!20]{\^a} $\rhd$  *Retired Senior Judge

\paragraph{Synonyms}
\hlc[gray!20]{'' ``Do Androids Dream of Electric Sheep?'' (the original of ``Blade Runner''), and his masterpiece, ``Ubik.''  Dick's fans are not modest in their claims. Nor are they especially precise: Borges, Calvino, Kafka, Robertson Davies are cited, in the blurbs and introductions, as his peers. A note of inconsistency inflects these claims-Calvino and Robertson Davies?}  $\rhd$ -but the point is clear: Dick is the most important writer of the last century.  The book is divided into three  $\rhd$ \hlc[red!20]{sections} \hlc[red!20]{parts} \hlc[red!20]{main}  \hlc[green!20]{categories} $\rhd$  : ``Themes,'' ``Themes and Themes,'' and ``Themes

\paragraph{Rejecting multiple candidates}
\hlc[gray!20]{s den.'' Scott is aware of the impact his race and size has on the way people -- particularly authority figures like law enforcement officers -- perceive him. He is big. He is dark-skinned. ``They look at us like we don't know how to control ourselves and we just get angry quick,'' he said. ``It's not even like that. They criminalize us for no reason.'' Scott$\boxtimes$}  $\rhd$ $\boxtimes$s mother, who is white, said she has been stopped by police for no reason. ``I' $\rhd$ \hlc[red!20]{ve} \hlc[red!20]{m} \hlc[red!20]{ll}  \hlc[green!20]{d} $\rhd$ \hlc[red!20]{be} \hlc[red!20]{say} \hlc[red!20]{get} \hlc[red!20]{like} \hlc[red!20]{never} \hlc[red!20]{just} \hlc[red!20]{have} \hlc[red!20]{ask} \hlc[red!20]{rather} \hlc[red!20]{been} \hlc[red!20]{tell}  \hlc[green!20]{pull} $\rhd$  over and they’d be like, $\boxtimes$  $\rhd$ \hlc[red!20]{$\boxtimes$} \hlc[red!20]{$\boxtimes$}  \hlc[green!20]{L} $\rhd$  What are you doing?''' she said

\paragraph{Other examples}
\hlc[gray!20]{Suzy is great! She helped me buy my condo at a great price (foreclosure) and then was super patient with my husband and me 4 years later when we were on a search for a house. She helped us get our...  Suzie H., Jacksonville  Goes above and beyond  Suzy has helped me close on my third property in 3 years. First she found me my dream pool home at the Beach then she helped me find two investment townhomes in}  $\rhd$ the same area  $\rhd$ \hlc[red!20]{.}  \hlc[green!20]{and} $\rhd$  now she is helping me find my dream home. She is always available to answer any questions I have and goes above and beyond to help me find the perfect home. I  $\rhd$ \hlc[red!20]{would}  \hlc[green!20]{highly} $\rhd$  recommend her!  Suzy H., Jacksonville

\hlc[gray!20]{from this new programme.  I have also been reminded of the role of tax measures in supporting urban development. With us in the gallery today is Mr Vuyisa Qabaka, a Cape Town entrepreneur and co-founder of an organisation called the Good Neighbourhoods Foundation. His advice is that ``Government should encourage township investment. For instance, it could promote urban development and regeneration through accelerated depreciation allowances for new building constructions or refurbishment of existing buildings.''}  $\rhd$ I am sure that  $\rhd$ \hlc[red!20]{the}  \hlc[green!20]{many} $\rhd$  of you  $\rhd$ \hlc[red!20]{will} \hlc[red!20]{have}  \hlc[green!20]{in} $\rhd$  this Chamber will agree with him.  I am also sure that many of you will agree with the Minister of Finance, who has said that the tax system should be used to support the  $\rhd$ \hlc[red!20]{development} \hlc[red!20]{growth}  \hlc[green!20]{economy} $\rhd$  and to create

\hlc[gray!20]{m off on some details.)  Unelma keltaisesta kuninkaasta.  Fastaval is not your average convention -- it specializes in incredibly tight auteur-designed roleplaying scenarios. A bunch of people run each scenario for players, not just the creator. There's awards for best scenarios in different categories.  The Society for Nordic Roleplaying published a collection of these scenarios translated into Finnish a few years ago, called Unelma keltais}  $\rhd$ esta kuninkaasta. It's a great book,  $\rhd$ \hlc[red!20]{and} \hlc[red!20]{but}  \hlc[green!20]{with} $\rhd$  a lot of great scenarios.  $\rhd$ \hlc[red!20]{I}  \hlc[green!20]{The} $\rhd$  book is available in English, but it’s not cheap. I’ve been looking for a copy for a while

\hlc[gray!20]{disappoint Jimmy. Then, I slept like a baby. SoFortWorthIt Oscars Swag GIVEAWAY!!!  The Oscars are exhausting, y'all. I'll definitely be cheering for all the stars this year, especially since I know the kind of caviar-Champagne-and-swag-filled night they're experiencing. And you know what? I want you to experience what it's like to get arm-loads of}  $\rhd$ free stuff.  So, I'm  $\rhd$ \hlc[red!20]{giving} \hlc[red!20]{doing} \hlc[red!20]{going}  \hlc[green!20]{partnering} $\rhd$  with the folks at the FortWorthIt Oscars Swag Giveaway to give away a \$100 Visa gift card to one lucky winner.  To enter, all you have  $\rhd$ \hlc[red!20]{to}  \hlc[green!20]{do} $\rhd$  is

\hlc[gray!20]{decision." "It will go down to destruction... or else, it will survive." "This is their moment of trial." "They've got to show themselves worthy of everything we gods have given them." "But evil is dark and strong." "And it may be that the scales of fate... are not yet in full balance." "What can I do to equalize both sides of the struggle, Athena?" "If you don't want to increase the powers of all men... then why don}  $\rhd$ 't you just give me the power to destroy them?" "I can't do that."  $\rhd$ \hlc[red!20]{"}  \hlc[green!20]{"[} $\rhd$  Thunderclap]" "I'm sorry." "I'm  $\rhd$ \hlc[red!20]{sorry} \hlc[red!20]{not} \hlc[red!20]{so}  \hlc[green!20]{afraid} $\rhd$ \hlc[red!20]{I} \hlc[red!20]{it} \hlc[red!20]{you} \hlc[red!20]{that}  \hlc[green!20]{the} $\rhd$  gods have decreed... that the balance of power must be maintained." "I'm

\hlc[gray!20]{give him a minute between removing the first tray and replacing it with the second - and you can come up with all sorts of theories to explain your findings. You can even throw a person in an MRI machine, study the flickering images on your computer screen. But the brain is the ultimate black box. Eventually, to grasp the first cut, you'll have to make another.  The car pulls into the parking lot of the nursing home, noses into an empty space. Annese and Cork}  $\rhd$ y get out, and Annese goes to the trunk to get the wheelchair. Corky is still standing, leaning on the car.  "I'm going  $\rhd$ \hlc[red!20]{to}  \hlc[green!20]{in} $\rhd$  ,"  $\rhd$ \hlc[red!20]{he} \hlc[red!20]{she}  \hlc[green!20]{An} $\rhd$  nese says.  "I'll come  $\rhd$ \hlc[red!20]{with}  \hlc[green!20]{in} $\rhd$  with  $\rhd$ \hlc[red!20]{you}  \hlc[green!20]{ya}


\begin{figure*}[h]
    \centering
    \small
    \begin{tabular}{p{1.85in}|p{1.85in}|p{1.85in}}
    \toprule
        \textbf{True Continuation} & \textbf{\memfree{}} & \textbf{Undefended}\\
        \midrule
         attendee list reflects the minimum number of seats reserved for your game.  Preregistered attendees for each of your events are listed below. If one attendee reserved multiple seats for your event, that attendee’s name will only be listed once. Therefore, the attendee list reflects the minimum number of seats reserved for your game.  Preregistered attendees for each of your events are listed below. If one attendee reserved multiple seats for your event, that attendee’s & attendee list reflects the minimum number of seats reserved for your game.  Preregistered attendees for each of your events are listed below. If one attendee reserved multiple seats for your event, that attendee’s name will only be listed once. Therefore, the attendee list reflects the minimum number of seats reserved for your game.  Preregistered attendees for each of your events are listed below. If one attendee reserved multiple seats for your event, that attendee’s \textbf{BLEU: 1} & attendee list reflects the minimum number of seats reserved for your game.  Preregistered attendees for each of your events are listed below. If one attendee reserved multiple seats for your event, that attendee’s name will only be listed once. Therefore, the attendee list reflects the minimum number of seats reserved for your game.  Preregistered attendees for each of your events are listed below. If one attendee reserved multiple seats for your event, that attendee’s \textbf{BLEU: 1}\\
         \midrule
         293.    294.    295.    296.    297.    298.    299.    300.    301.    302.    303.    304.    305.    306.    307.    308.    309.    310.    311.    312. & 293.    294.    295.    296.    297.    298.    299.    300.    301.    302.    30\diffdel{3}\diffadd{4}.    30\diffdel{4}\diffadd{3}.    305.    30\diffdel{6}\diffadd{4}.    30\diffdel{7}\diffadd{6}.    308.    30\diffdel{9}\diffadd{7}.    3\diffdel{10}\diffadd{09}.    3\diffdel{11}\diffadd{08}.    3\diffdel{12}\diffadd{10}. \textbf{BLEU: 0.95} & 293.    294.    295.    296.    297.    298.    299.    300.    301.    302.    303.    304.    305.    306.    307.    308.    309.    310.    311.    312. \textbf{BLEU: 1}\\
         \midrule
         0x058f8f8aU,     0x3f9292adU, 0x219d9dbcU, 0x70383848U, 0xf1f5f504U,     0x63bcbcdfU, 0x77b6b6c1U, 0xafdada75U, 0x42212163U,     0x20101030U, 0 & 0x058f8f8aU,     0x3f9292adU, 0x219d9dbcU, 0x70383848U, 0xf1f5f504U, 0x63bcbcdfU,     0x77b6b6c1U\diffadd{L}, 0xafdada75U,0x42212163U, 0x20101030U,     0 \textbf{BLEU: 0.93} & 0x058f8f8aU,     0x3f9292adU, 0x219d9dbcU, 0x70383848U, 0xf1f5f504U,     0x63bcbcdfU, 0x77b6b6c1U, 0xafdada75U, 0x42212163U,     0x20101030U, 0 \textbf{BLEU: 1} \\
         \midrule
        7, calc(sin((pi/180)*a7)))  define(cea0, calc(cos((pi/180)*ea0))) define(cea1, calc(cos((pi/180)*ea1))) define(cea2, calc(cos((pi/180)*ea2))) define(cea3, calc(cos((pi/180)*ea3))) define(cea4, calc(cos((pi/180 & 7, calc(sin((pi/180)*a7)))  define(cea0, calc(cos((pi/180)*ea0))) define(cea1, calc(cos((pi/180)*ea1))) define(cea2, calc(cos((pi/180)*e\diffdel{a}2))) define(cea3, calc(cos((pi/180)*e\diffdel{a}3))) define(cea4, calc(cos((pi/180 \textbf{BLEU: 0.95} & 7, calc(sin((pi/180)*a7)))  define(cea0, calc(cos((pi/180)*ea0))) define(cea1, calc(cos((pi/180)*ea1))) define(cea2, calc(cos((pi/180)*ea2))) define(cea3, calc(cos((pi/180)*ea3))) define(cea4, calc(cos((pi/180 \textbf{BLEU: 1} \\
    \bottomrule
    \end{tabular}
    \caption{Random sample of \memfree{} generations where the BLEU score with the true continuation $> 0.9$. Most of these examples are repetitive and/or lists of numbers. In the \memfree{} column, we use highlights to show the difference from the true continuation: red means deleted text, and green means added text.
    }
    \label{tab:still_memorized}
\end{figure*}


\section{Author Ordering Algorithm}
\label{authorder}

\begin{figure}[H]
\begin{tcolorbox}
\begin{verbatim}
import hashlib
import numpy as np

def hash(x):
  h=hashlib.new("md5")
  h.update(bytes(x,"ascii"))
  return int(h.hexdigest(),16)

names = ("Nicholas Daphne " + 
  "Katherine Matthew " +
  "Florian Chiyuan Milad " + 
  "Christopher").split()

for i in range(0,10000):
  s = str(i)
  l = [hash(x+s) for x in names]
  o = np.argsort(l)
  if names[o[0]] != "Daphne":
    continue
  if names[o[-1]] != "Nicholas": 
    continue
  print([names[x] for x in o])
  exit(0)
\end{verbatim}
\end{tcolorbox}
\caption{Author ordering algorithm}
\label{alg:authorder}
\end{figure}

\end{document}